\documentclass[preprint,12pt]{elsarticle}

\usepackage{amssymb}
\usepackage{amsmath}

\usepackage{algorithm}
\usepackage{algorithmic}
\usepackage{subfig}
\usepackage{multirow}
\usepackage{booktabs}
\usepackage{longtable}
\usepackage{bm}

\journal{Applied Soft Computing}

\begin{document}

\begin{frontmatter}



\title{Impact of Surrogate Model Accuracy on Performance and Model Management Strategy in Surrogate-Assisted Evolutionary Algorithms} 

\author[label1]{Yuki Hanawa\corref{cor1}}
\ead{hanawa-yuki@ed.tmu.ac.jp}
\cortext[cor1]{Corresponding author}
\author[label2]{Tomohiro Harada}
\ead{tharada@mail.saitama-u.ac.jp}
\author[label3]{Yukiya Miura}
\ead{miura@tmu.ac.jp}
\affiliation[label1]{organization={Graduate School of Systems Design, Tokyo Metropolitan University},
 addressline={Asahigaoka 6-6},
 city={Hino},
 postcode={191-0065},
 state={Tokyo},
 country={Japan}}

\affiliation[label2]{organization={Graduate School of Science and Engineering, Saitama University},
 addressline={255, Sakura-ku Shimo-okubo},
 city={Saitama},
 postcode={338-8570},
 state={Saitama},
 country={Japan}}

\affiliation[label3]{organization={Faculty of Systems Design, Tokyo Metropolitan University},
 addressline={Asahigaoka 6-6},
 city={Hino},
 postcode={191-0065},
 state={Tokyo},
 country={Japan}} 



\begin{abstract}
Surrogate-assisted evolutionary algorithms (SAEAs) have been proposed to solve expensive optimization problems. Although SAEAs use surrogate models that approximate the evaluations of solutions using machine learning techniques, prior research has not adequately investigated the impact of surrogate model accuracy on search performance and model management strategy in SAEAs. This study analyzes how surrogate model accuracy affects search performance and model management strategies. For this purpose, we construct a pseudo-surrogate model with adjustable prediction accuracy to ensure fair comparisons across different model management strategies. We compared three model management strategies: (1) pre-selection (PS), (2) individual-based (IB), and (3) generation-based (GB) on standard benchmark problems with a baseline model that does not use surrogates. The experimental results reveal that a higher surrogate model accuracy improves the search performance. However, the impact varies according to the strategy used. Specifically, PS demonstrates a clear trend of improved performance as the estimation accuracy increases, whereas IB and GB exhibit robust performance when the accuracy surpasses a certain threshold. In model strategy comparisons, GB exhibits superior performance across a broad range of prediction accuracies, IB outperforms it at lower accuracies, and PS outperforms it at higher accuracies. The findings of this study clarify guidelines for selecting appropriate model management strategies based on the surrogate model accuracy.

\end{abstract}




\end{frontmatter}



\section{Introduction}
Optimization is the process of maximizing or minimizing an objective (evaluation) function under specific constraints. Optimization problems are expressed using the following mathematical formula:
\begin{align*}
\text{minimize} \quad & f(\bm{x})\\
\text{subject to} \quad & g_i(\bm{x})\leq 0 \quad (i=1, \ldots, n)\\
& h_j(\bm{x}) = 0 \quad (j=1, \ldots, p)
\end{align*}
where $\bm{x}$ represents the design variables, $f$ is the objective function, and $g_i$ and $h_j$ are the inequality and equality constraint functions, respectively, and $n$ and $p$ indicate the number of each type of constraint.

Evolutionary algorithms (EAs) are a class of metaheuristics that solve optimization problems by mimicking the mechanism of natural evolution to search for optimal solutions~\cite{MEZURAMONTES2011173,Slowik2020}. EAs have demonstrated high search performance, particularly in high-dimensional spaces and problems with complex constraints. However, a challenge arises when the computational cost of the evaluation function is high (e.g., in problems requiring simulations or complex numerical computations). These problems are known as expensive optimization problems (EOPs)~\cite{EOP}, which require extensive computation time because of the large number of evaluations required during the optimization process. 

Surrogate-assisted evolutionary algorithms (SAEAs) ~\cite{HE2023119495,JIN201161,Jin2021} have attracted attention for addressing this challenge. SAEAs replace expensive solution evaluations with low-cost approximation models (surrogate models) using machine learning (ML) techniques. By employing surrogate models, SAEAs could achieve acceptable optimization results within a realistic execution time, leveraging low-cost approximated evaluations with surrogate models to explore solutions based on past histories.

Owing to the use of surrogate models, their prediction accuracy of surrogate models affects the search performance of SAEAs \cite{9240452}. Specifically, an accurate surrogate model allows for more precise predictions of evaluation values, reduces the number of actual evaluation functions, and shortens the overall execution time. However, less accurate surrogate models might lead to erroneous search directions, yet they could enhance search performance by facilitating escape from the local optima.

From this perspective, the research questions addressed in this study are:
\begin{description}
 \item[RQ1:] How does the prediction accuracy of surrogate models impact the search performance of SAEAs?
 \item[RQ2:] Do different SAEA model management strategies vary in their sensitivity to surrogate model accuracy?
 \item[RQ3:] Could the optimal SAEA model management strategy be determined based on surrogate model accuracy?
\end{description}

To answer these research questions, we constructed a pseudo-surrogate model with an adjustable prediction accuracy using actual evaluation functions. We compared the search performance of SAEAs using pseudo-surrogate models with varying accuracies. 

This study addresses three SAEA model management strategies: (1) pre-selection (PS), (2) individual-based (IB), and (3) generation-based (GB). We compared the combinations of these three strategies and pseudo-surrogate models of varying accuracy on CEC 2015 benchmark problems \cite{Chen2015ProblemDA} to evaluate the search performance under a limited evaluation budget. In this study, we aim to provide guidelines for selecting appropriate model management strategies based on surrogate model accuracy, thereby enhancing the optimization performance of SAEAs.

In our previous study, \cite{10611759} conducted a preliminary analysis of the relationship between the prediction accuracy of surrogate models and the SAEA strategies. However, this early work used only two strategies (PS and IB) with different surrogate representations, which restricted the direct comparison of model management strategies under consistent accuracy metrics. In contrast, this study incorporated a new SAEA strategy, GB, in addition to PS and IB, and used a common pseudo-surrogate model. This improvement enabled consistent comparisons across all strategies, allowing us to answer the research questions more precisely and comprehensively.

\section{SAEAs}
Over the past several decades, SAEAs employing various surrogate models have been proposed~\cite{HE2023119495,JIN201161,Jin2021}. This section provides an overview of SAEAs. Then, we explain three representative model management strategies: PS, IB, and GB, based on the classifications provided in previous studies \cite{TONG2021414,Jin2005}.

\subsection{Overview of SAEAs}
SAEA is a method that reduces computational costs by using surrogate models, which are generally constructed using ML techniques (e.g., the radial basis function, Kriging model, and neural networks). Specifically, during the optimization process, SAEA constructs a surrogate model from the information on solution evaluations obtained from the past search history. Then, SAEA uses the constructed surrogate model to predict the fitness values of the solutions and determine the solutions to be evaluated using an actual expensive evaluation function. Only solutions predicted to be promising undergo a detailed evaluation with the actual evaluation function; otherwise, less promising solutions are discarded. This approach enables the omission of actual evaluations for less promising solutions, thereby reducing the overall computational costs.

\subsection{PS SAEA}
The PS strategy uses a surrogate model to select superior individuals before performing actual evaluations. In general, the PS strategy generates several offspring. Then, the surrogate model selects the most promising individual. Subsequently, the selected individuals undergo an actual evaluation. This approach allows superior individuals to be retained more easily, thereby reducing computational costs. Examples of SAEAs using PS include SACCDE \cite{YANG202050} and CHDE+ELDR \cite{10001982} for constrained optimization and KTA2 \cite{9406113} and MCEA/D \cite{9733428} for multi-objective optimization.

PS could employ several types of surrogate models, including absolute, rank-based, and classification-based surrogates. In a PS with an absolute evaluation surrogate model, multiple offspring candidates are generated, and their evaluation values are predicted. Then, the superior solutions are preselected and evaluated. In a PS with a rank-based surrogate model, the rankings of the generated offspring are predicted, and those with higher ranks are preselected for evaluation. In PS with a classification-based surrogate model, the superiority or inferiority of the solutions compared to their parent models is learned. If the offspring are predicted to be superior, they are evaluated; otherwise, they are rejected without actual evaluation. If the actual evaluation is better than that of the parent, the parent replaces it; otherwise, the parent remains in the next generation.

\begin{algorithm}[tb]
 \caption{A pseudocode of the pre-selection strategy SAEA (PS) used in this study \cite{TONG2021414}}
 \label{algorithm:original-PS-CS}
 \begin{algorithmic}[1]
 \STATE Create the initial sample of \(5 \times d\) individuals using Latin Hypercube Sampling (LHS) ~\cite{McKay1979LHS}
 \STATE Evaluate all initial individuals and store them in an archive $\mathcal{A}$
 \STATE Set\:\(FE=5\times d\)
 \STATE Select the best $N$ individuals from $\mathcal{A}$ for the population $P$
 \WHILE{\( FE < maxFE \)}
 \STATE Perform crossover and mutation operators to generate $N$ offspring
 \FOR{each offspring $off$}
 \STATE Find its parent $par$ as a reference individual
 \STATE Build a surrogate $\mathcal{S}$ using $\mathcal{A}$
 \STATE Predict a label whether $off$ is better than $par$ using $\mathcal{S}$
 \IF{predictive label is true}
 \STATE Evaluate the offspring with an actual evaluation function
 \STATE Add all actual evaluated offspring to $\mathcal{A}$
 \STATE \(FE = FE + 1\)
 \IF{$off$ is superior to $par$}
 \STATE Replace $par$ with $off$ in the population $P$
 \ENDIF
 \ENDIF
 \ENDFOR
 \ENDWHILE
 \end{algorithmic}
\end{algorithm}
Algorithm~\ref{algorithm:original-PS-CS} presents the pseudocode of PS with the classification-based surrogate model employed in this study. Where $d$ denoted the dimensions of the design variables, $N$ denoted population size, and $maxFE$ denoted the maximum number of evaluations. All evaluated individuals were stored in an archive $\mathcal{A}$ and used for surrogate construction. In PS, offspring were generated through crossover and mutation operators (Line 6). Subsequently, a classification-based surrogate model $\mathcal{S}$ was constructed for each offspring using its parent as a reference individual (Lines 8--9). Then, the surrogate model predicted the superiority of the offspring over its parent individual (Line 10). If it was predicted to be superior, the offspring was evaluated using the actual evaluation function (Line 12). After the actual evaluation was confirmed, it was compared with its parent and replaced if it was superior to its parent (Lines 14--16). Conversely, if offspring were predicted to be inferior, they were rejected without applying an actual evaluation function.

\subsection{IB SAEA}
The IB strategy uses a surrogate model to determine whether each individual offspring should undergo actual evaluation. Generally, the IB strategy selects offspring that are predicted to be the most promising based on the evaluation values predicted by the surrogate model. Then, the selected offspring were evaluated using an actual evaluation function. This approach reduces the number of evaluations by excluding individuals who are unlikely to perform well. Examples of SAEAs that use IB include GPEME \cite{6466380}, VESAEA \cite{8789910}, and RFMOISR \cite{10.1007/978-3-540-24694-7_73}. 

IB could be combined with absolute fitness and rank-based surrogates. In IB with an absolute fitness surrogate model, the evaluation values of the newly generated offspring are predicted, and the offspring with higher predicted values are evaluated. Finally, the next population was selected based on actual and predicted evaluation values. IB with a rank surrogate model predicts ranking within the offspring population and performs actual evaluations for higher-ranked individuals until the correlation with the training data reaches a certain level. Once sufficient correlation is achieved, the top-ranked individuals are selected as the next generation's population.

\begin{algorithm}[tb]
 \caption{A pseudocode of the individual-based strategy (IB) used in this study \cite{TONG2021414}}
 \label{algorithm:original-IB-AFM}
 \begin{algorithmic}[1]
 \STATE Create the initial sample of \(5 \times d\) individuals using Latin Hypercube Sampling (LHS)
 \STATE Evaluate all initial individuals and store them in an archive $\mathcal{A}$
 \STATE Set\:\(FE=5\times d\)
 \STATE Select the best $N$ individuals from $\mathcal{A}$ for the population $P$
 \WHILE{\(FE < maxFE\)}
 \STATE Perform crossover and mutation operators to generate $N$ offspring
 \STATE Build a surrogate $\mathcal{S}$ using $\mathcal{A}$
 \STATE Sort the parent and offspring individuals using $\mathcal{S}$
 \STATE Evaluate \(r_{sm}\times N\) best predicted individuals with actual function
 \STATE \(FE = FE + r_{sm} \times N\)
 \STATE Select $N$ best individuals from parents and offspring for the next generation
 \ENDWHILE
 \end{algorithmic}
\end{algorithm}
Algorithm~\ref{algorithm:original-IB-AFM} presents the pseudocode for IB used in this study. In IB, offspring are generated through crossover and mutation and are sorted using a surrogate model (Line 8). Then, the top $r_{sm}\times N\;(0<r_{sm}<1)$ individuals are re-evaluated using the actual evaluation function, and the top $N$ individuals from both parents and offspring are selected for the next generation's population (Lines 8--11). During this process, unevaluated individuals were compared based on their predicted evaluation values.

\subsection{GB SAEA}

The GB strategy~\cite{JIN201161} evolved the population using only a surrogate model for specific generations, without using an actual evaluation function during this period. After completing a set number of generations, the most promising individual in the final population was evaluated using the evaluation function. Examples of SAEAs using GB include SAHO~\cite{PAN2021304}, LSA-FIDE~\cite{YU2024120246}, and GORS-SSLPSO~\cite{YU201914}. GB could be combined with two types of surrogates: absolute fitness- and rank-based. With either type of surrogate, the search continues for a certain number of generations while performing parent and survival selection based on surrogate-based fitness values.

\begin{algorithm}[tb]
 \caption{A pseudocode of the generation-based strategy (GB) used in this study ~\cite{Jin2021}}
 \label{algorithm:original-GB}
 \begin{algorithmic}[1]
 \STATE Generate the initial sample of \(5 \times d\) individuals using Latin Hypercube Sampling (LHS)
 \STATE Evaluate all initial individuals and store them in an archive $\mathcal{A}$
 \STATE Set\:\(FE=5\times d\)
 \STATE Select the best $N$ individuals from $\mathcal{A}$ for the population $P$
 \WHILE{\(FE < maxFE\)}
 \STATE \(gen = 0\)
 \WHILE{\(gen < maxGen\)}
 \STATE Perform crossover and mutation operators to generate $N$ offspring
 \STATE Build a surrogate $\mathcal{S}$ using $\mathcal{A}$
 \STATE Sort the parent and offspring individuals using $\mathcal{S}$ 
 \STATE \(gen = gen + 1\)
 \STATE Select $N$ best individuals from parents and offspring for the next generation
 \ENDWHILE
 \STATE Evaluate best predicted individuals with an actual function
 \STATE Add an actual evaluated individual to $\mathcal{A}$
 \STATE \(FE = FE + 1\)
 \STATE Select $N$ best individuals from parents and offspring for the next generation
 \ENDWHILE
 \end{algorithmic}
\end{algorithm}

Algorithm \ref{algorithm:original-GB} shows the pseudocode for GB, which is addressed in this study. Here, $gen$ represents the generation count within the loop, and $maxGen$ represents the number of generations required to evolve the population using only a surrogate model. In GB, candidate solutions are evaluated for a certain number of generations using only the surrogate while repeating crossover and mutation operators (Lines 7--13). After a specific number of generations, the individual with the best fitness value predicted by the surrogate model underwent an actual evaluation, and the evaluation count was updated (Lines 14--16). The surrogate was retrained using an archive that included the newly evaluated individual.

\section{Pseudo-surrogate Models}

To enable performance analysis of SAEAs using surrogate models with arbitrary prediction accuracy, this study proposes a pseudo-surrogate model. The pseudo-surrogate model determines its predictions by directly using actual evaluation values rather than by learning from past history. The pseudo-surrogate model allows for adjustable prediction accuracy. To enable adjustment of surrogate model accuracy in PS, IB, and GB, the actual evaluation function was used directly to replicate the surrogate models artificially. Note that for all models, the actual evaluations used in the pseudo-surrogate are not included in the number of actual evaluations ($FE$) within the algorithm.

The pseudocode for the pseudo-surrogate model is presented in Algorithm~\ref{algorithm:PS-CM}. In Algorithm~\ref{algorithm:PS-CM}, $sp$ indicated the surrogate prediction accuracy and was an adjustable parameter. $rand(0, 1)$ returned a random value in the range $[0, 1]$. In the pseudo-surrogate model, two solutions, $x_1$ and $x_2$, were evaluated and compared (Lines 1--2). First, the actual superiority was set to the variable $label$ and reversed with a certain probability of ($1-sp$) to artificially replicate the prediction error (Lines 3--5).

\begin{algorithm}[tb]
\caption{A pseudo-surrogate model to compare two individuals $x_1$ and $x_2$}
\label{algorithm:PS-CM}
\begin{algorithmic}[1]
\STATE Evaluate $x_1$ and $x_2$ with an actual evaluation function
\STATE $label \gets (f(x_1) < f(x_2))$
\IF{$rand(0,1) < (1 - sp)$}
 \STATE Flip $label$
\ENDIF
\STATE \textbf{return} $label$
\end{algorithmic}
\end{algorithm}

In the PS strategy, the pseudo-surrogate model shown in Algorithm~\ref{algorithm:PS-CM} is used in Lines 9--10 of Algorithm~\ref{algorithm:original-PS-CS}, which simply replaces a surrogate model $\mathcal{S}$.

For the IB and GB strategies, a pseudo-surrogate model was used to sort the populations. Algorithm~\ref{algorithm:bubble_sort} is a sorting algorithm that uses a pseudo-surrogate model. In Algorithm~\ref{algorithm:bubble_sort}, $P$ denoted the population to be sorted. This sorting algorithm was based on bubble sorting, in which a comparison of two individuals was performed as follows:
\begin{itemize}
 \item If both individuals were actually evaluated, their actual evaluations were compared without error (Lines 4--7).
 \item If either individual was unevaluated, they were compared based on the predictions of the pseudo-surrogate model with a prediction accuracy of $sp$ (Lines 8--12).
\end{itemize}
Note that the actual evaluation values used in this sort exclude those evaluated specifically for the pseudo-surrogate model and refer only to those obtained during the algorithm execution procedure. Algorithm~\ref{algorithm:bubble_sort} is used in Lines 7--8 of Algorithm~\ref{algorithm:original-IB-AFM} for IB, while it is used in Lines 9--10 of Algorithm~\ref{algorithm:original-GB} for GB.

The use of this pseudo-surrogate model enables the reproduction of SAEA behavior with a surrogate model of arbitrary accuracy. In addition, it ensures fair comparisons between model management strategies because all strategies use the same pseudo-surrogate model (see Algorithm~\ref{algorithm:original-GB}).

\begin{algorithm}[tb]
\caption{Sorting algorithm with the pseudo-surrogate model}
\label{algorithm:bubble_sort}
\begin{algorithmic}[1]
\STATE $n = |P|$
\FOR{$i = 1$ to $n$}
\FOR{$j = 1$ to $n - i$}
\IF{Both individuals evaluated with an actual fitness function}
\IF{\(f(P_j)>f(P_{j+1})\)}
\STATE swap $P_j$ and $P_{j+1}$ 
\ENDIF
\ELSE
\IF{The pseudo-surrogate model (Algorithm~\ref{algorithm:PS-CM}) returns false}
\STATE swap $P_j$ and $P_{j+1}$
\ENDIF
\ENDIF
\ENDFOR
\ENDFOR
\end{algorithmic}
\end{algorithm}

\section{Experiments}
To investigate the impact of the surrogate model accuracy on the search performance of SAEAs, we conducted experiments using a pseudo-surrogate model. The following subsection explains the compared SAEA strategies and presents the experimental setup. Finally, the benchmark functions used in the experiment are detailed.

\subsection{Compared Strategies}
For the SAEA methods, this experiment employed three model management strategies: PS, IB, and GB, with the pseudo-surrogate model explained in the previous section. In addition, a method without a surrogate model (NoS) was compared as a baseline. Algorithm~\ref{algorithm:NoS} presented the pseudocode for the NoS. The flow of the NoS differed based on the compared algorithms. When comparing NoS with PS (Lines 7--14), all the generated offspring were always evaluated with an actual evaluation. If an offspring was superior to its parent, it replaced the parent individual for use in the next generation's population. However, when comparing NoS with IB and GB (Lines 16--19), all generated offspring were evaluated using an actual evaluation function, and the top $N$ individuals from both parents and offspring were selected for the next generation. In addition to not using a surrogate, other procedures were the same as those used for PS, IB, and GB.
\begin{algorithm}[tb]
\caption{A pseudocode for a method without a surrogate model (NoS) used in the experiment \cite{TONG2021414}}
\label{algorithm:NoS}
\begin{algorithmic}[1]
\STATE Create the initial sample of \(5 \times d\) individuals using Latin Hypercube Sampling (LHS)
\STATE Evaluate all initial individuals
\STATE Set\:\(FE=5\times d\)
\STATE Select the best $N$ individuals from the sample for the population $P$
\WHILE{\( FE < \text{maxFE} \)}
 \STATE Perform crossover and mutation operators to generate $N$ offspring
 \STATE \textbf{/***PS***/}
 \FOR{each offspring $off$}
 \STATE Find its parent $par$ as the reference individual
 \STATE Evaluate the offspring
 \IF{\(f(off) < f(par)\)}
 \STATE Replace $par$ with $off$ in $P$
 \ENDIF
 \ENDFOR
 \STATE
 \STATE \textbf{/***IB and GB***/}
 \STATE Evaluate all offspring
 \STATE Select $N$ best individuals from parents and offspring for the next generation
\ENDWHILE
\end{algorithmic}
\end{algorithm}

\subsection{Test Problems}
This experiment used a variety of test problems from the CEC2015 competition \cite{Chen2015ProblemDA}; in particular, $f1$, $f2$, $f4$, $f8$, $f13$, and $f15$ were used. Table~\ref{tb:details_of_benchmark_and_optimal_value} presents the fitness landscapes, optimal values, and dimensions. The objective functions $f1$ and $f2$ were unimodal, $f4$ and $f8$ were multimodal, and $f13$ and $f15$ combined the features of both the unimodal and multimodal functions. In this experiment, the dimensionality of the design variables was set to 10 and 30, respectively.

\begin{table}[tb]
\caption{Details of benchmarks and optimal values \cite{TONG2021414}}
\label{tb:details_of_benchmark_and_optimal_value}
\begin{center}
\begin{tabular}{cccc}
\hline
\textbf{Problem} & \textbf{Landscape} & \textbf{Optimal Value ($f^*$)} & \textbf{Dimensions} \\
\hline
$f1$ & \multirow{2}{*}{Unimodal} & 100 & \multirow{6}{*}{10, 30} \\

$f2$ & & 200 & \\

$f4$ & \multirow{2}{*}{Simple Multimodal} & 400 & \\

$f8$ & & 800 & \\

$f13$ & \multirow{2}{*}{Composition Function} & 1300 & \\
$f15$ & & 1500 & \\
\hline
\end{tabular}
\end{center}
\end{table}

\subsection{Experimental Setup}
The population size was set to 40. The extended intermediate crossover (EIX) \cite{6792992} was employed as a crossover operator. First, the EIX selected two parent individuals from the population based on the crossover rate. Then, for each $i$-th variable of the offspring $t_i$, an arbitrary value $\alpha_i$ was randomly chosen between $\gamma$ and $1+\gamma$, and the $i$-th variable of the offspring was determined using the following formula:
\begin{equation}
 t_i = \alpha_i v_i + (1-\alpha_i) w_i
\end{equation}
where $v_i$ and $w_i$ represented the $i$th design variables of the two parent individuals. In this study, $\gamma$ was set to 0.4, in accordance with \cite{TONG2021414}. As a mutation operator, we used uniform mutation within the upper and lower limits of the design variables. The crossover rate $pc$ was set to 0.7, whereas the mutation rate $pm$ was set to $0.3$ following a previous study. For each method, the actual evaluation ratio $rsm$ for IB was set as 0.5, whereas $maxGen$ for GB was set as 30. The prediction accuracy of the pseudo-surrogate model was set as $sp={0.5, 0.6, 0.7, 0.8, 0.9, 1.0}$. When $sp=1.0$, no prediction error occurred, and all comparison results were always correct.

In these experiments, the maximum number of evaluations was limited to 2000, within which the progression of the best-found solution during the search process was analyzed. We conducted 21 independent trials for each problem and calculated their averages.

\section{Results and Discussion}
The experimental results are presented in this section. Section~\ref{sec:rel} analyzed the relationship between the search performance and prediction accuracy to answer RQ1 for each model management strategy. To answer RQ2, Section~\ref{sec:sen} compared the sensitivities of the three strategies to different surrogate model accuracies. Finally, to answer RQ3, Section~\ref{sec:com} compared the search performance across the three strategies for each prediction accuracy and revealed the optimal strategy for each accuracy. 

Experiments were conducted with dimensions of 10 and 30. However, the trends observed for the 10 and 30 dimensions were similar, and including both would make the study unnecessarily lengthy. Therefore, this study presented the results for 30 dimensions in the following subsection, owing to page limitations. The results for the 10 dimensions are provided in the Supplementary Material.

\subsection{Relation between Search Performance and Prediction Accuracy}\label{sec:rel}
First, we analyzed the relationship between search performance and prediction accuracy for PS, IB, and GB. 

Figures \ref{fig:pssvc_d30}--\ref{fig:gbrbf_d30} show the results of solving 30-dimensional problems using PS, IB, and GB, respectively. The horizontal axis represented the number of actual evaluations, whereas the vertical axis represented the difference between the optimal objective function value and that obtained using the algorithms. Different colors indicated the difference in the accuracy of the pseudo-surrogate model, whereas the black line indicated the result of the NoS. In the following subsections, we discuss the results of each model management strategy and conclude with an analysis of the correlation between the model accuracy and search performance.

\subsubsection{PS}

\begin{figure*}[tb] 
 \centering
 \subfloat[$f1$]{\includegraphics[bb = 0 0 720 432, width=0.32\linewidth]{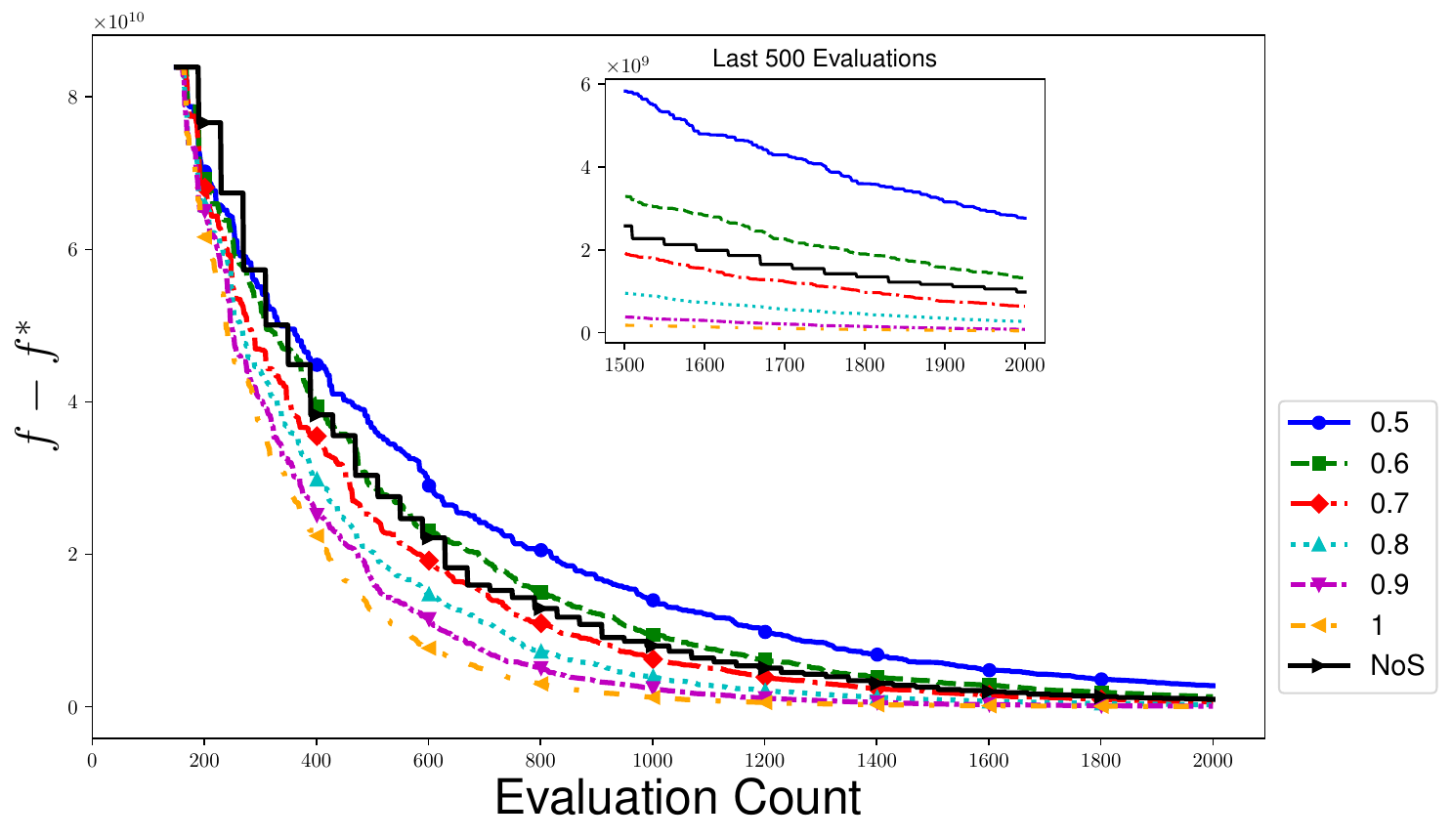}\label{fig:pssvc_f1_d30}}
 \hfill
 \subfloat[$f2$]{\includegraphics[bb = 0 0 720 432, width=0.32\linewidth]{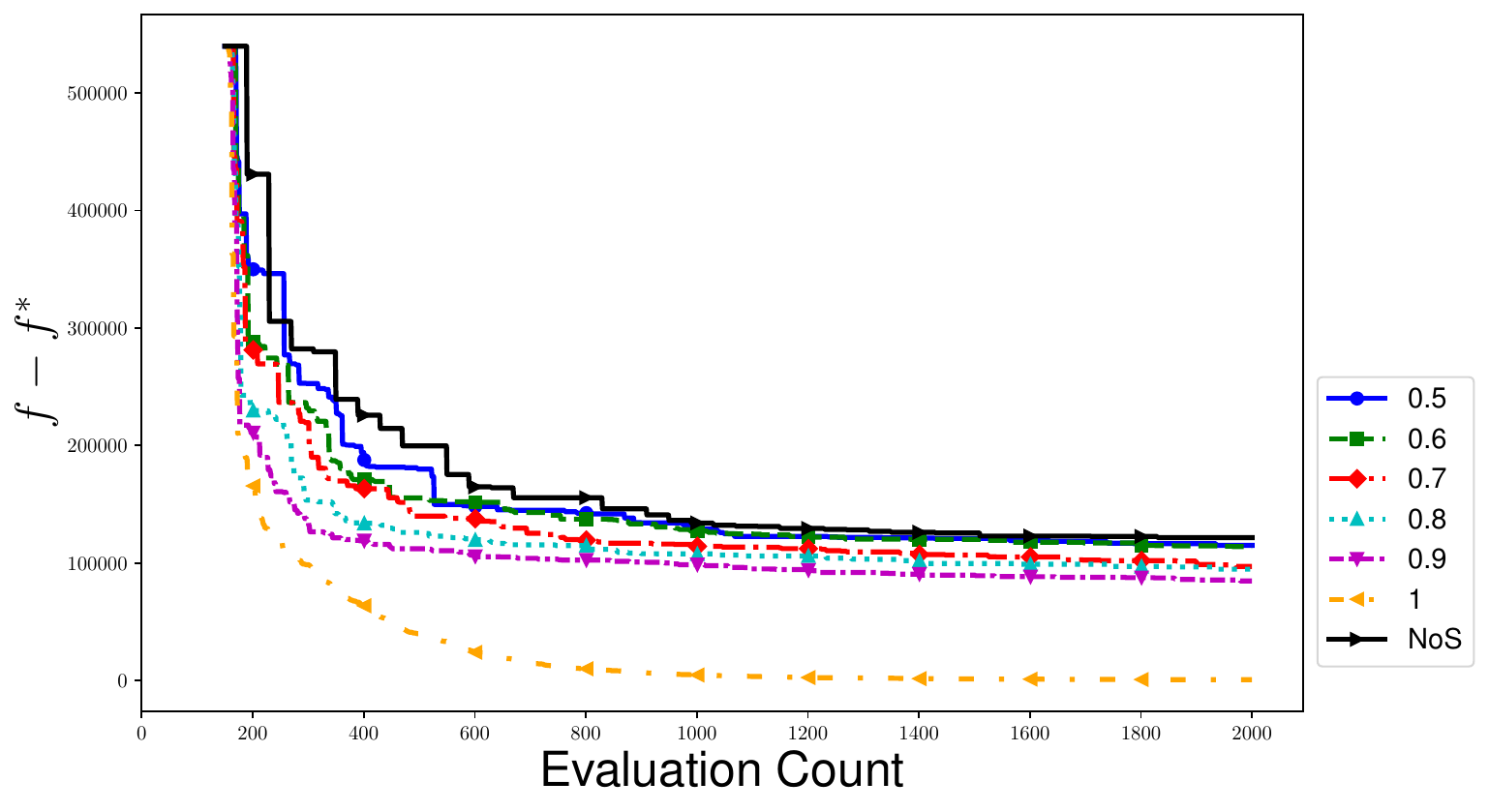}\label{fig:pssvc_f2_d30}}
 \hfill
 \subfloat[$f4$]{\includegraphics[bb = 0 0 720 432, width=0.32\linewidth]{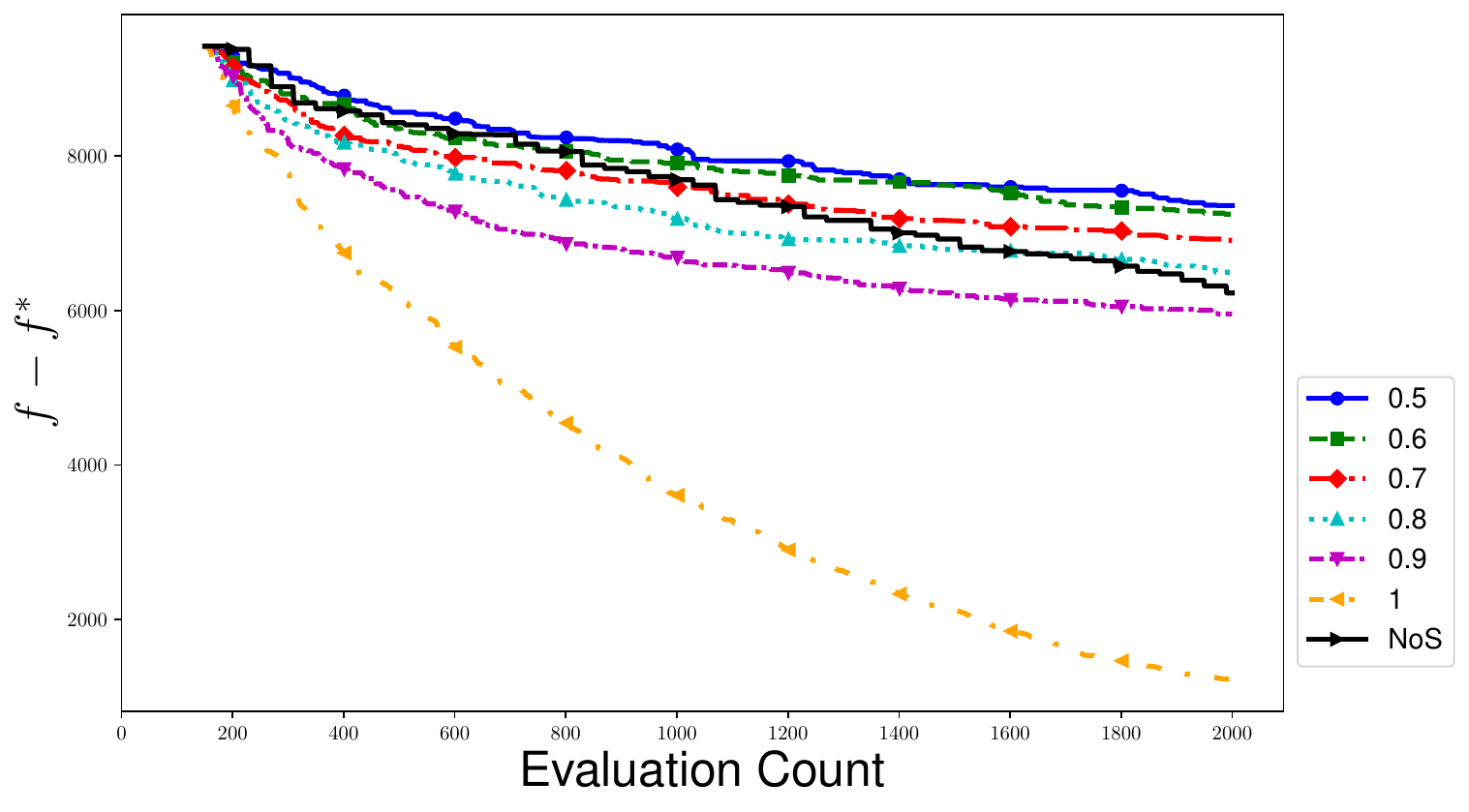}\label{fig:pssvc_f4_d30}}
 \\
 \subfloat[$f8$]{\includegraphics[bb = 0 0 720 432, width=0.32\linewidth]{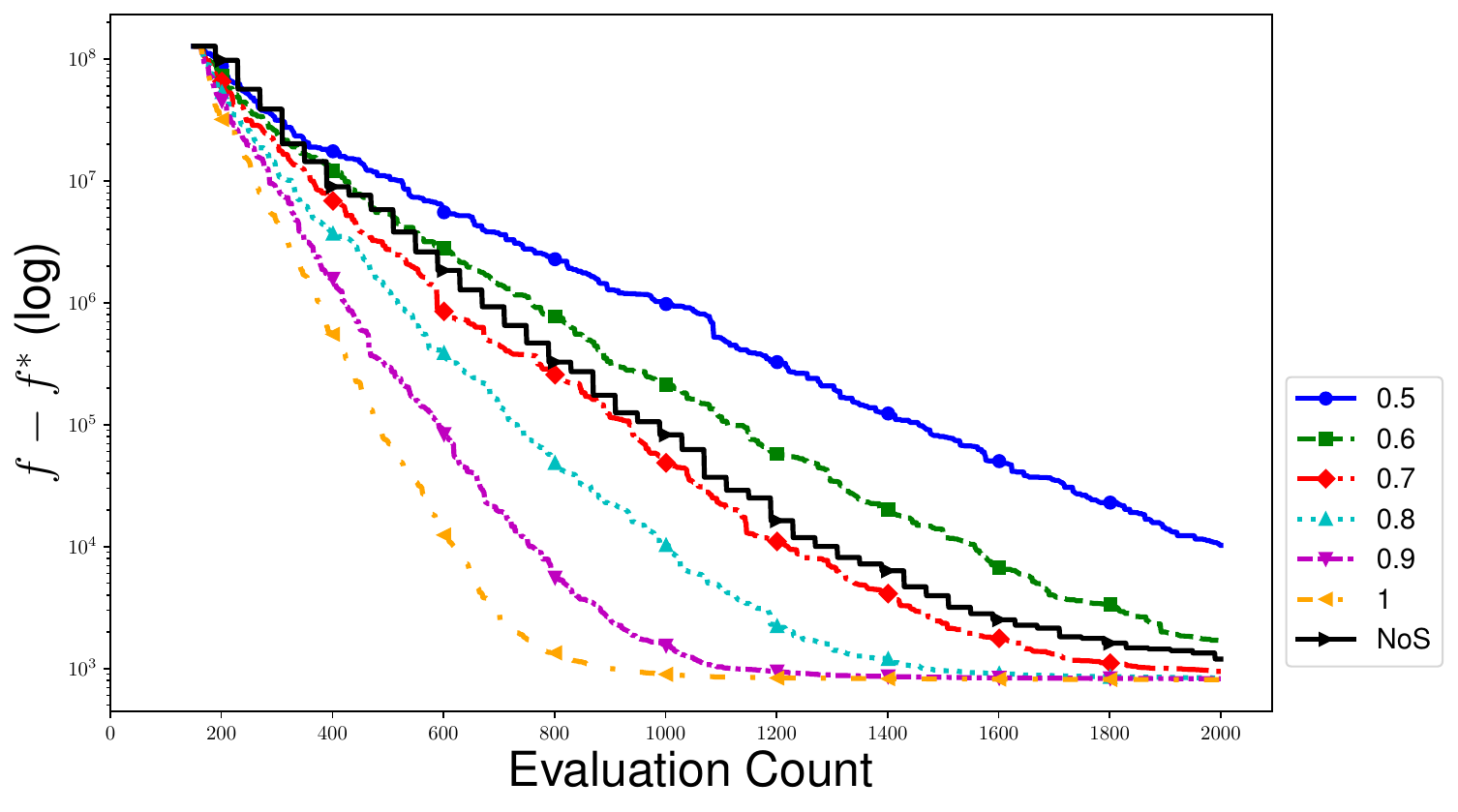}\label{fig:pssvc_f8_d30}}
 \hfill
 \subfloat[$f13$]{\includegraphics[bb = 0 0 720 432, width=0.32\linewidth]{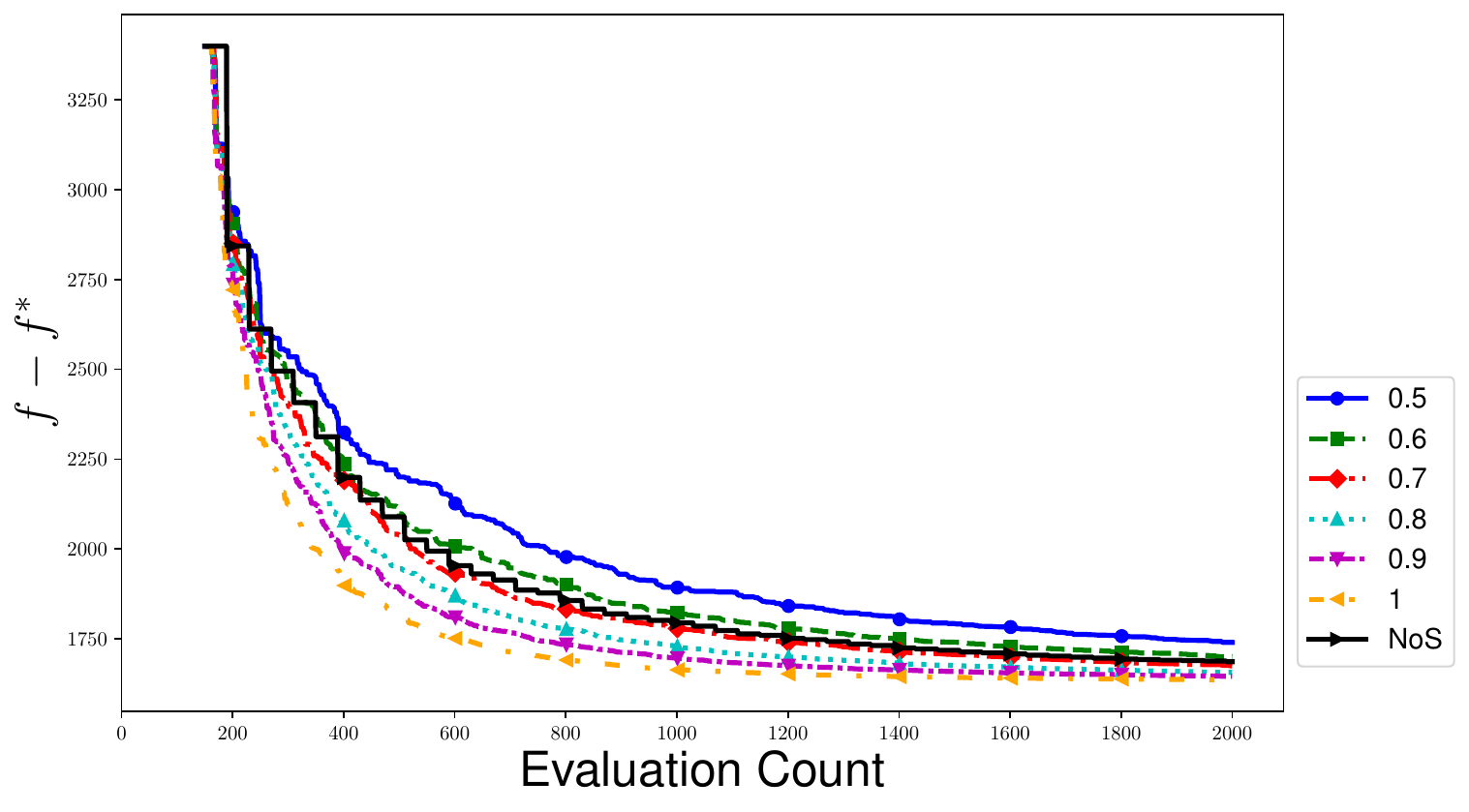}\label{fig:pssvc_f13_d30}}
 \hfill
 \subfloat[$f15$]{\includegraphics[bb = 0 0 720 432, width=0.32\linewidth]{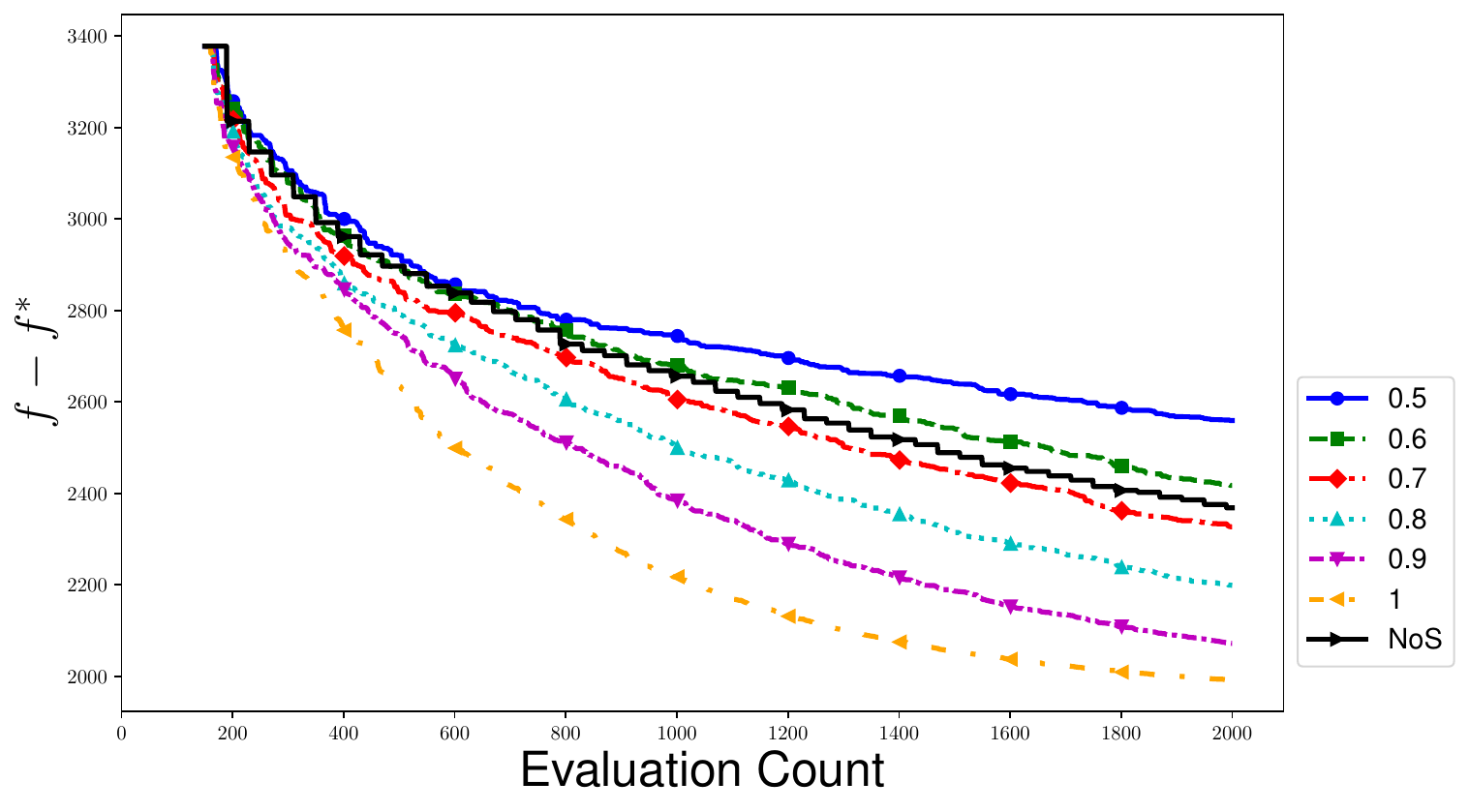}\label{fig:pssvc_f15_d30}}
 \caption{Transition of the difference between the optimal objective function value and the value obtained by algorithms when solving 30D problems using PS}
 \label{fig:pssvc_d30}
\end{figure*}

Figures \ref{fig:pssvc_f1_d30} and \ref{fig:pssvc_f2_d30} show the results for unimodal problems. For $f1$, the smallest objective function value was obtained with an accuracy of 1.0. In addition, after 200 evaluations, an accuracy of 1.0 constantly obtained the smallest objective function value. In addition, an accuracy of 0.5 was comparable to that of NoS, resulting in the worst search performance. Similarly, for $f2$, the search performance was the best, with an accuracy of 1.0. In addition, the accuracies of 0.5 and 0.6 were comparable to those of the NoS, resulting in the worst search performance.

Figures \ref{fig:pssvc_f4_d30} and \ref{fig:pssvc_f8_d30} show the results for multimodal problems. For $f4$, an accuracy of 1.0 achieved the smallest objective function value, and this trend was observed throughout the entire search process. In addition, an accuracy of 0.5 was comparable to that of the NoS, resulting in the poorest search performance. For $f8$, an accuracy of 1.0 achieved the smallest objective function value. In addition, an accuracy of 0.5 resulted in the lowest search performance.

Figures \ref{fig:pssvc_f13_d30} and \ref{fig:pssvc_f15_d30} present the composite function results. Similar to the other functions, for both $f13$ and $f15$, an accuracy of 1.0 achieved the smallest objective function value. In addition, an accuracy of 0.5, which was comparable to that of the NoS, led to the worst search results.

\subsubsection{IB}
\begin{figure*}[tb] 
 \centering
 \subfloat[$f1$]{\includegraphics[bb = 0 0 720 432, width=0.32\linewidth]{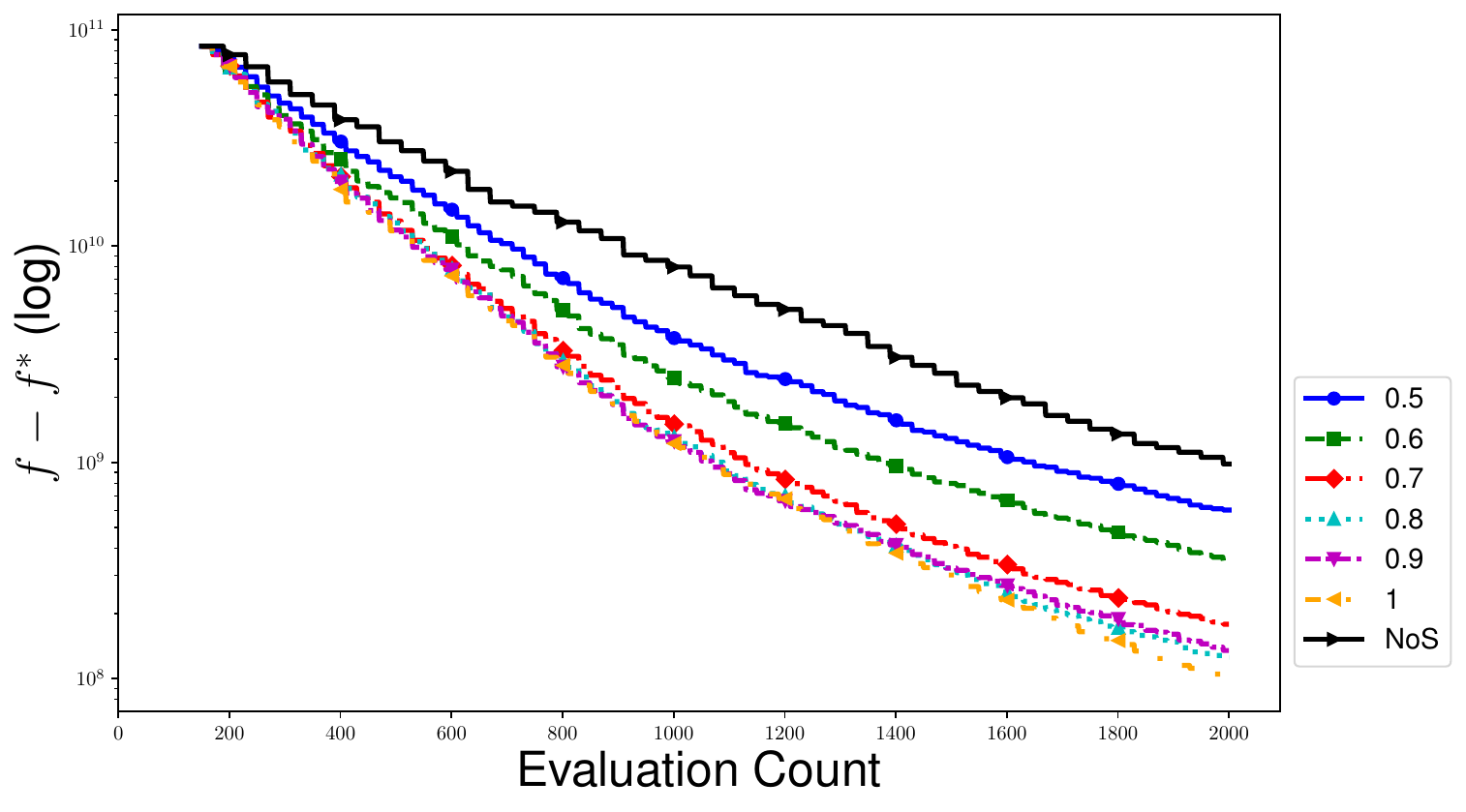}\label{fig:ibrbf_f1_d30}}
 \hfill
 \subfloat[$f2$]{\includegraphics[bb = 0 0 720 432, width=0.32\linewidth]{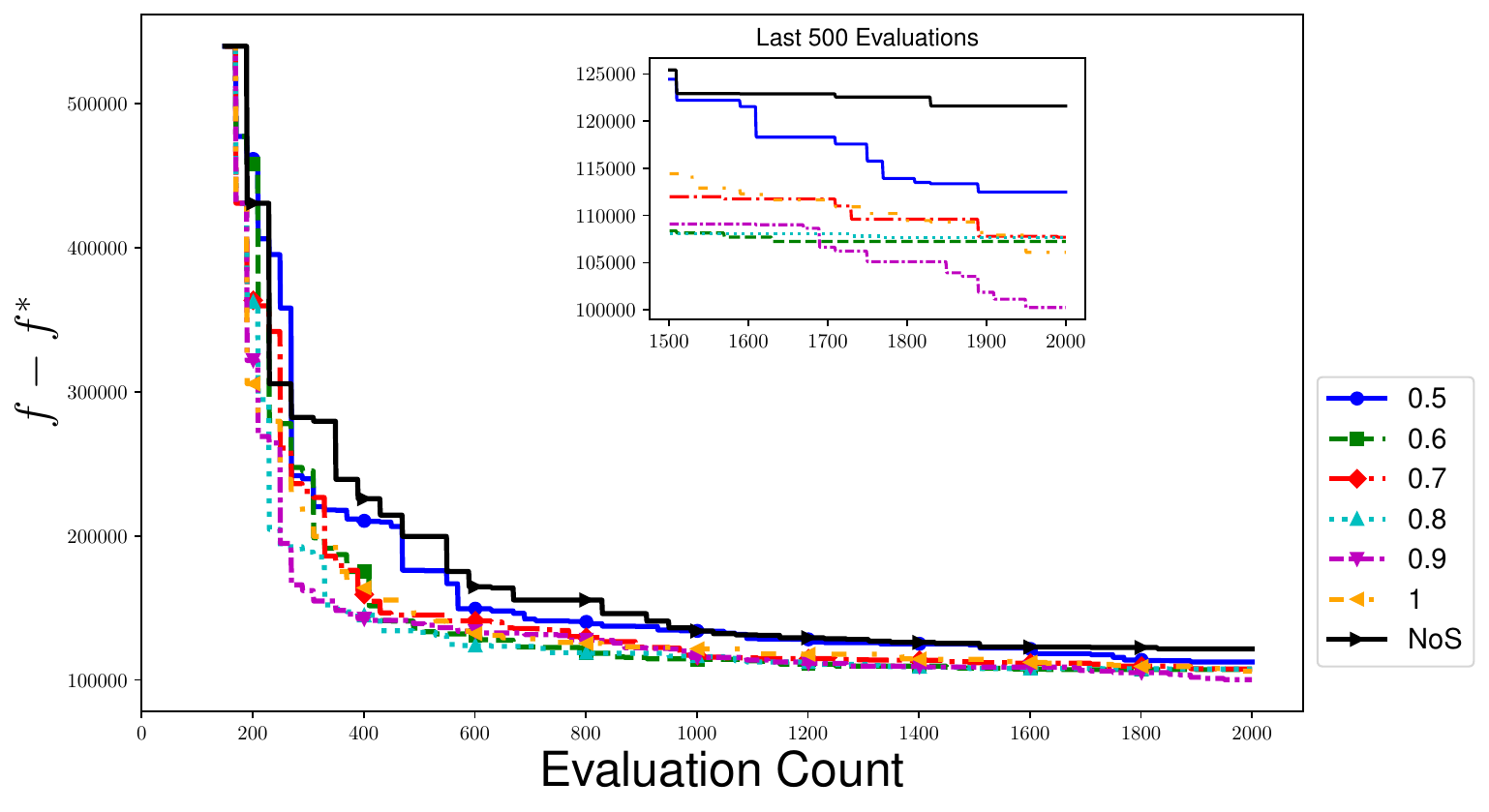}\label{fig:ibrbf_f2_d30}}
 \hfill
 \subfloat[$f4$]{\includegraphics[bb = 0 0 720 432, width=0.32\linewidth]{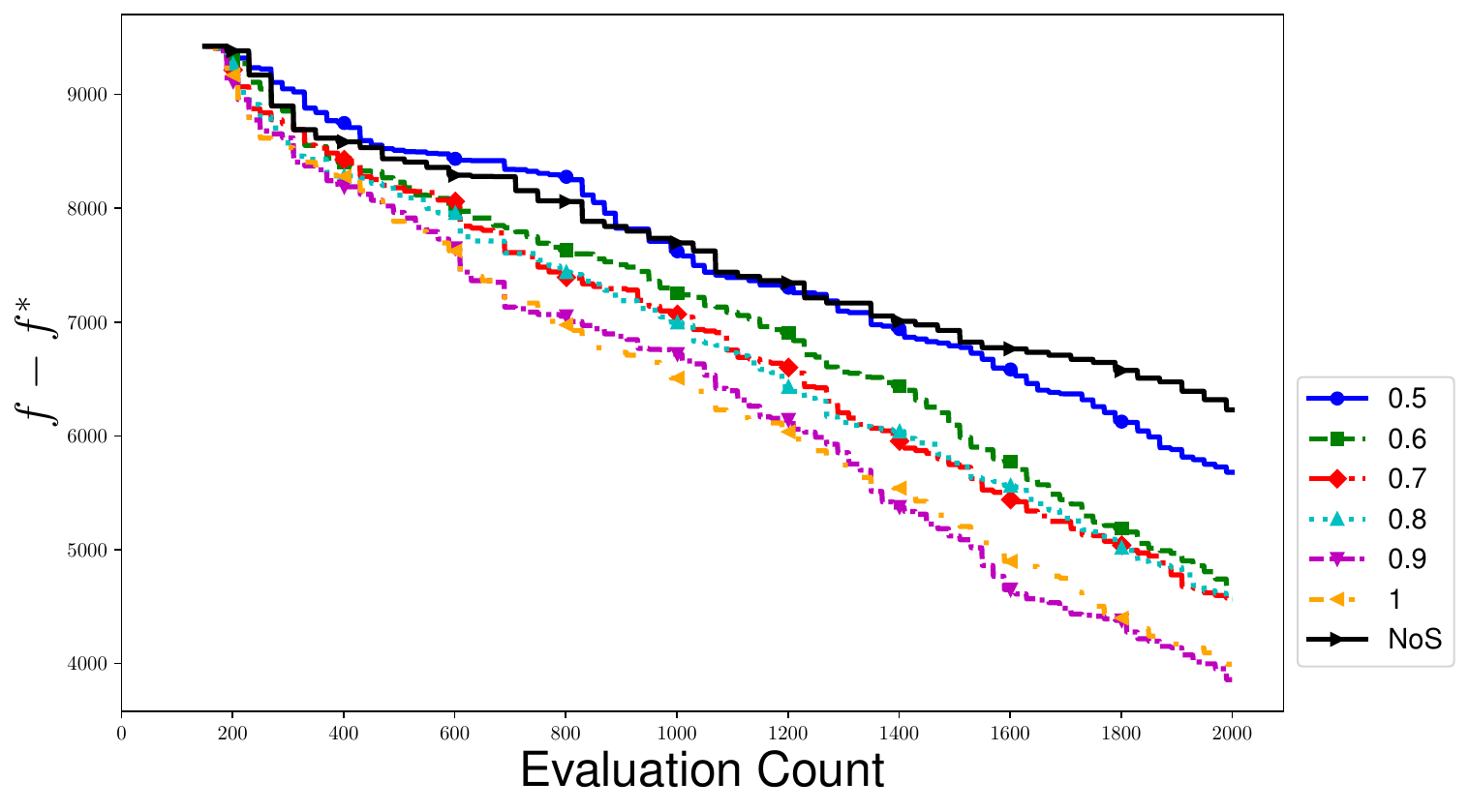}\label{fig:ibrbf_f4_d30}}
 \\
 \subfloat[$f8$]{\includegraphics[bb = 0 0 720 432, width=0.32\linewidth]{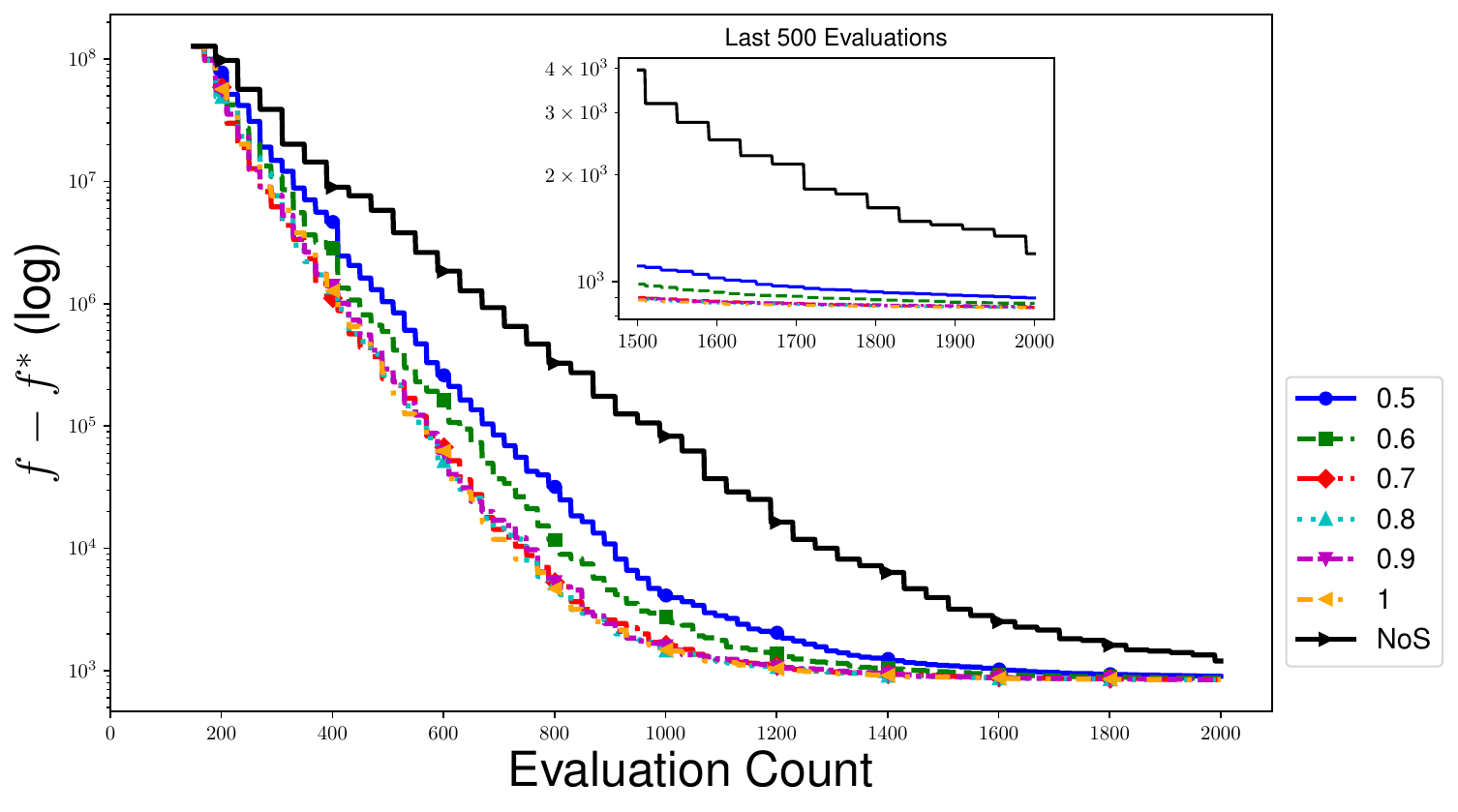}\label{fig:ibrbf_f8_d30}}
 \hfill
 \subfloat[$f13$]{\includegraphics[bb = 0 0 720 432, width=0.32\linewidth]{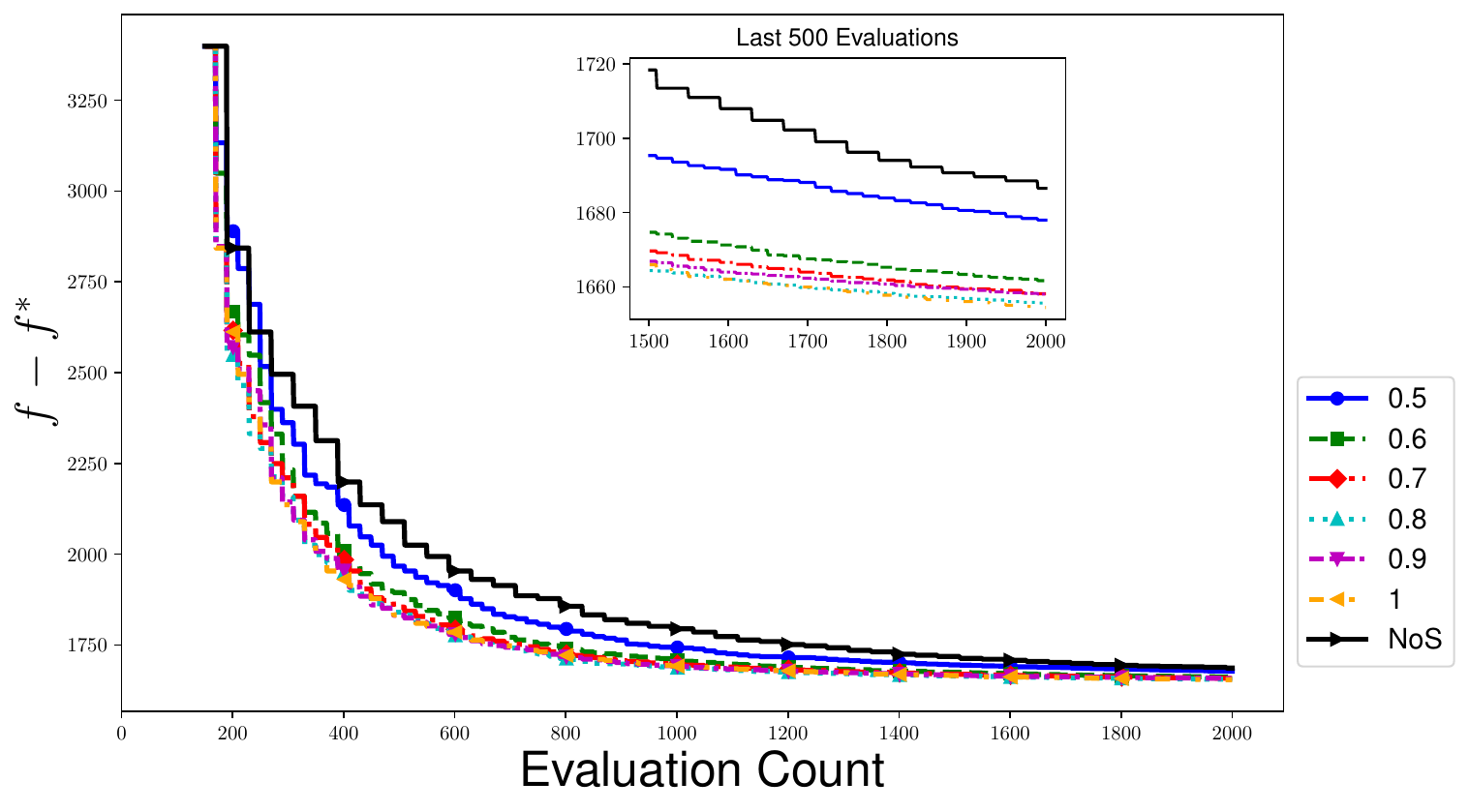}\label{fig:ibrbf_f13_d30}}
 \hfill
 \subfloat[$f15$]{\includegraphics[bb = 0 0 720 432, width=0.32\linewidth]{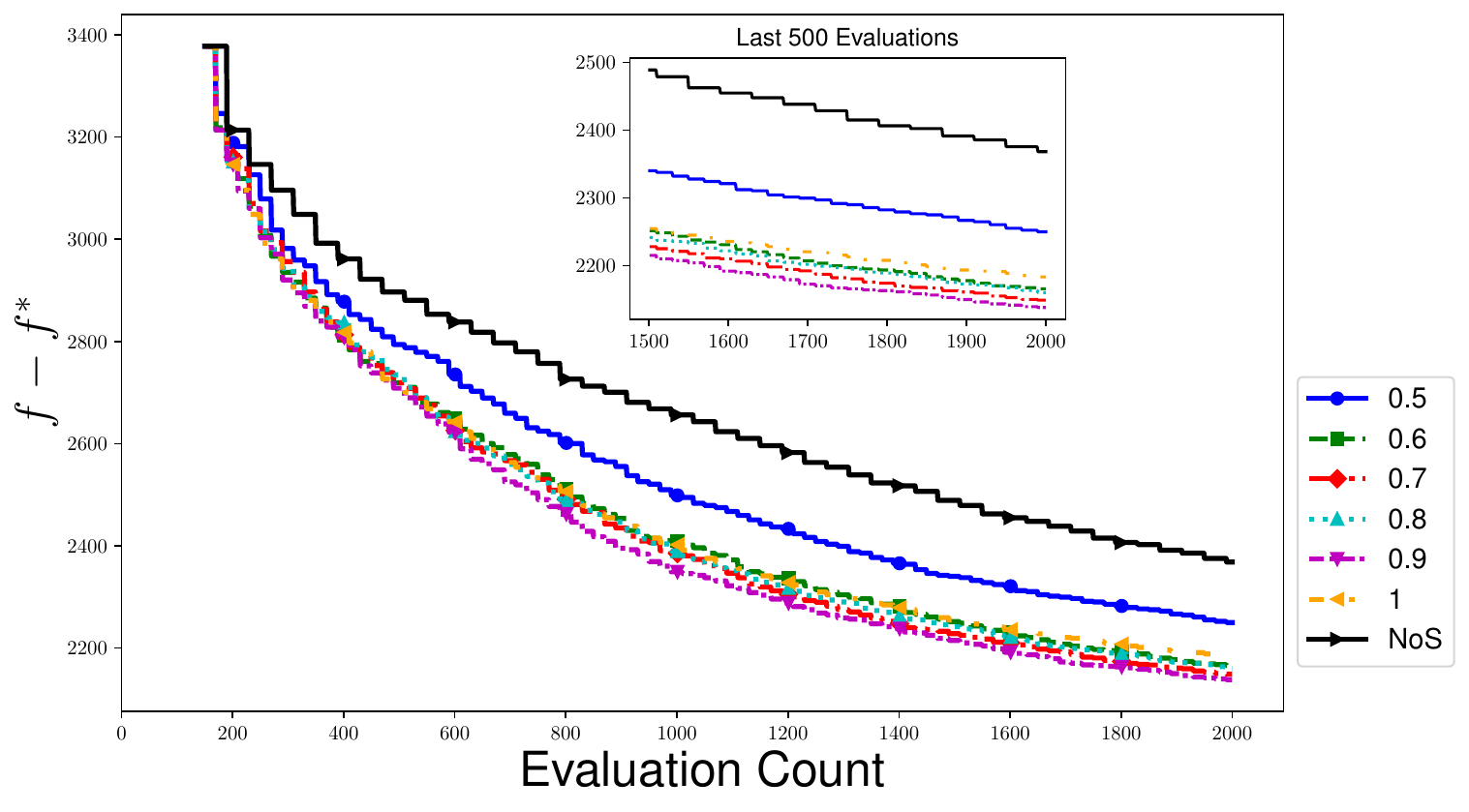}\label{fig:ibrbf_f15_d30}}
 \caption{Transition of the difference between the optimal objective function value and the value obtained by algorithms when solving 30D problems using IB}
 \label{fig:ibrbf_d30}
\end{figure*}

Figures \ref{fig:ibrbf_f1_d30} and \ref{fig:ibrbf_f2_d30} show the results for unimodal problems. For $f1$, the smallest objective function value was obtained with an accuracy of 1.0. In addition, with the exception of an accuracy of 0.8, higher accuracy indicated better search performance.
However, for $f2$, the best performance was obtained with an accuracy of 0.9. However, an accuracy of 1.0

Figures \ref{fig:ibrbf_f4_d30} and \ref{fig:ibrbf_f8_d30} show the results for multimodal problems. For $f4$, an accuracy of 0.9 achieved the smallest objective function value, and this trend was observed throughout the entire search process. NoS exhibited the worst search performance. With a surrogate model, an accuracy of 0.5 showed the worst search performance. For $f8$, an accuracy of 1.0 achieved the smallest objective function value. In contrast, the NoS exhibited the worst search performance, followed by an accuracy of 0.5.

Figures \ref{fig:ibrbf_f13_d30} and \ref{fig:ibrbf_f15_d30} present the composite function results. For $f13$, an accuracy of 1.0 achieved the smallest objective function value, followed by an accuracy of 0.8. NoS exhibited the worst search performance. For $f15$, accuracies of 0.9 and 0.7 achieved small objective function values in that order. In contrast, NoS demonstrated the poorest search performance, with an accuracy of 0.5 being better but still suboptimal.

\subsubsection{GB}
\begin{figure*}[tb] 
 \centering
 \subfloat[$f1$]{\includegraphics[bb = 0 0 720 432, width=0.32\linewidth]{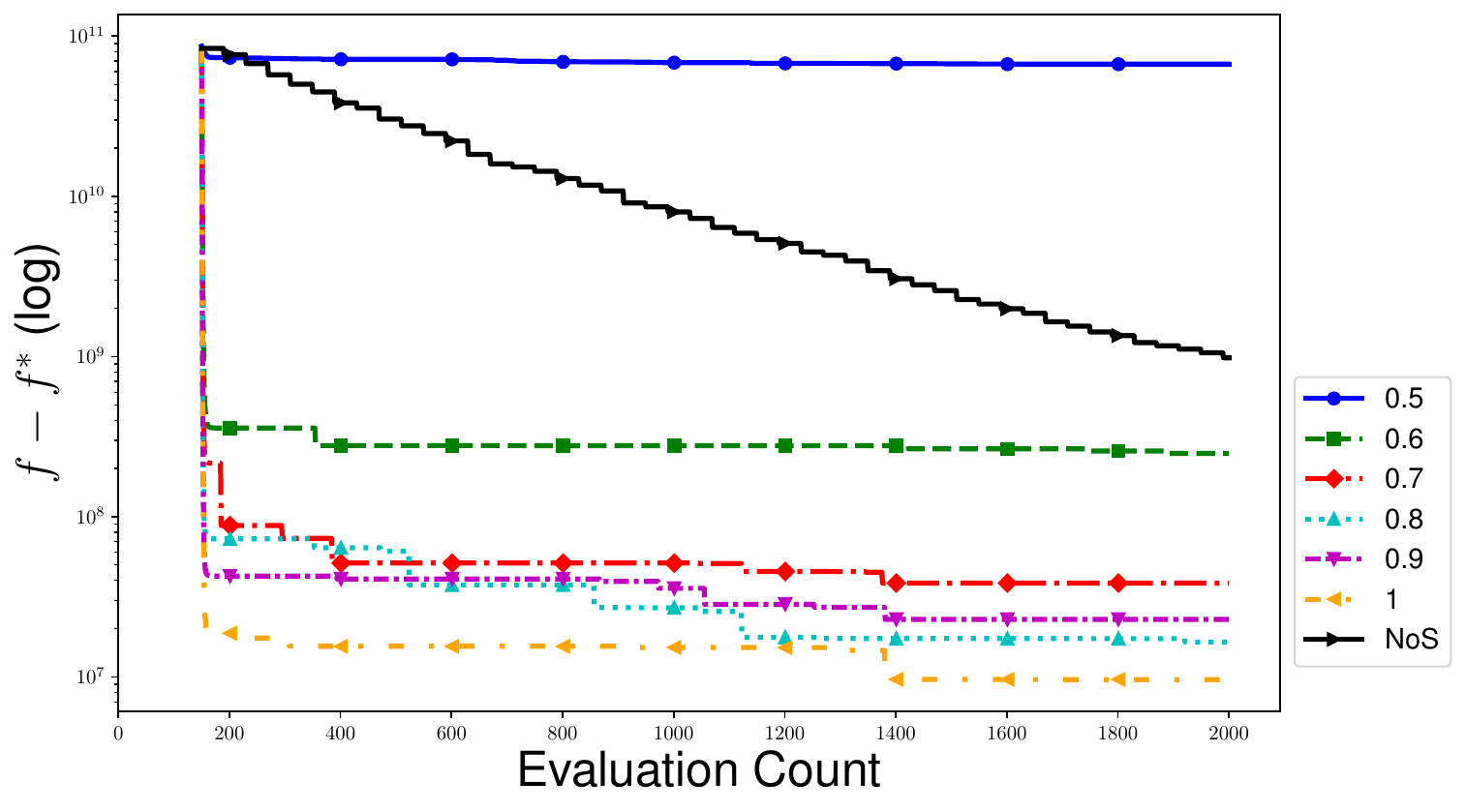}\label{fig:gbrbf_f1_d30}}
 \hfill
 \subfloat[$f2$]{\includegraphics[bb = 0 0 720 432, width=0.32\linewidth]{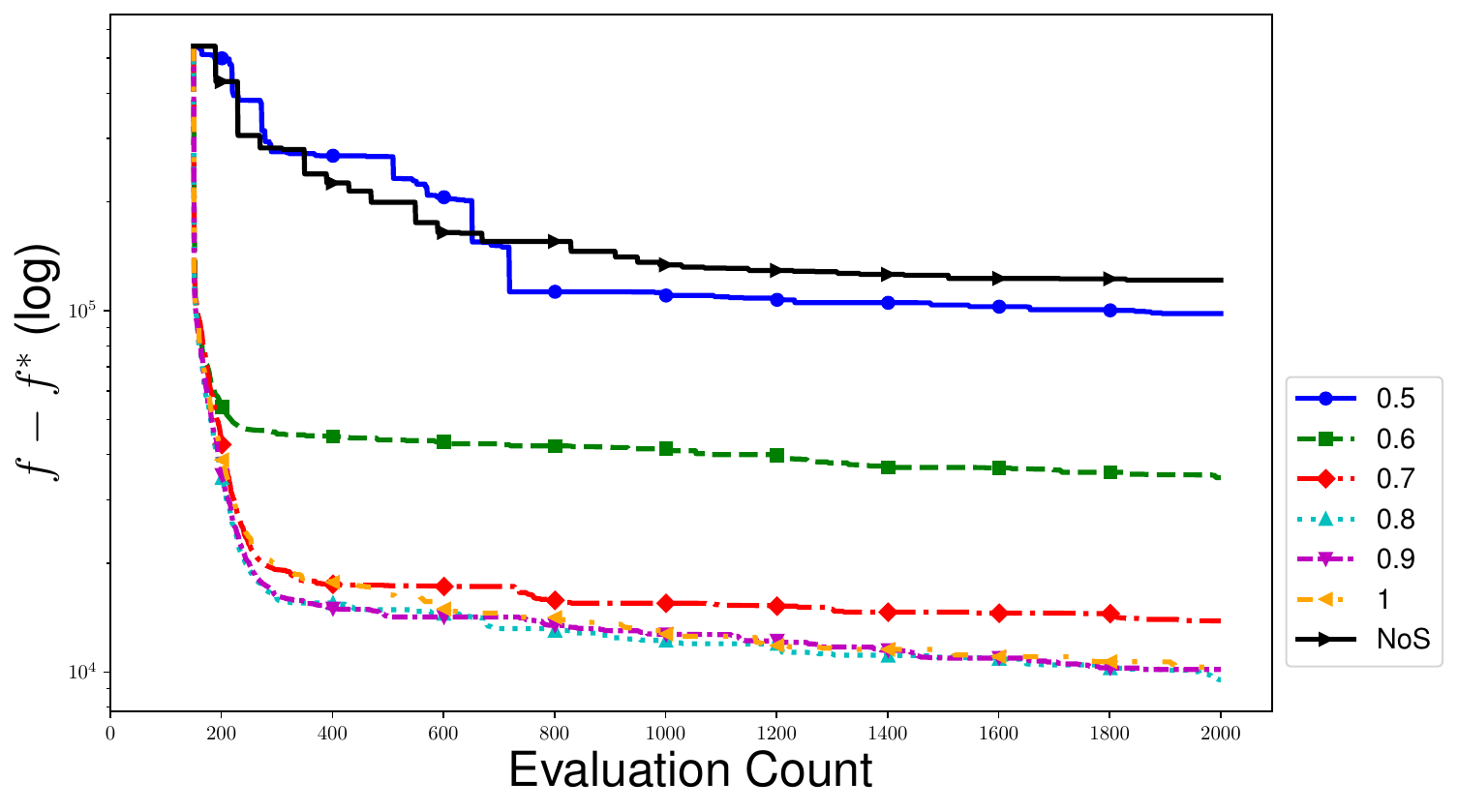}\label{fig:gbrbf_f2_d30}}
 \hfill
 \subfloat[$f4$]{\includegraphics[bb = 0 0 720 432, width=0.32\linewidth]{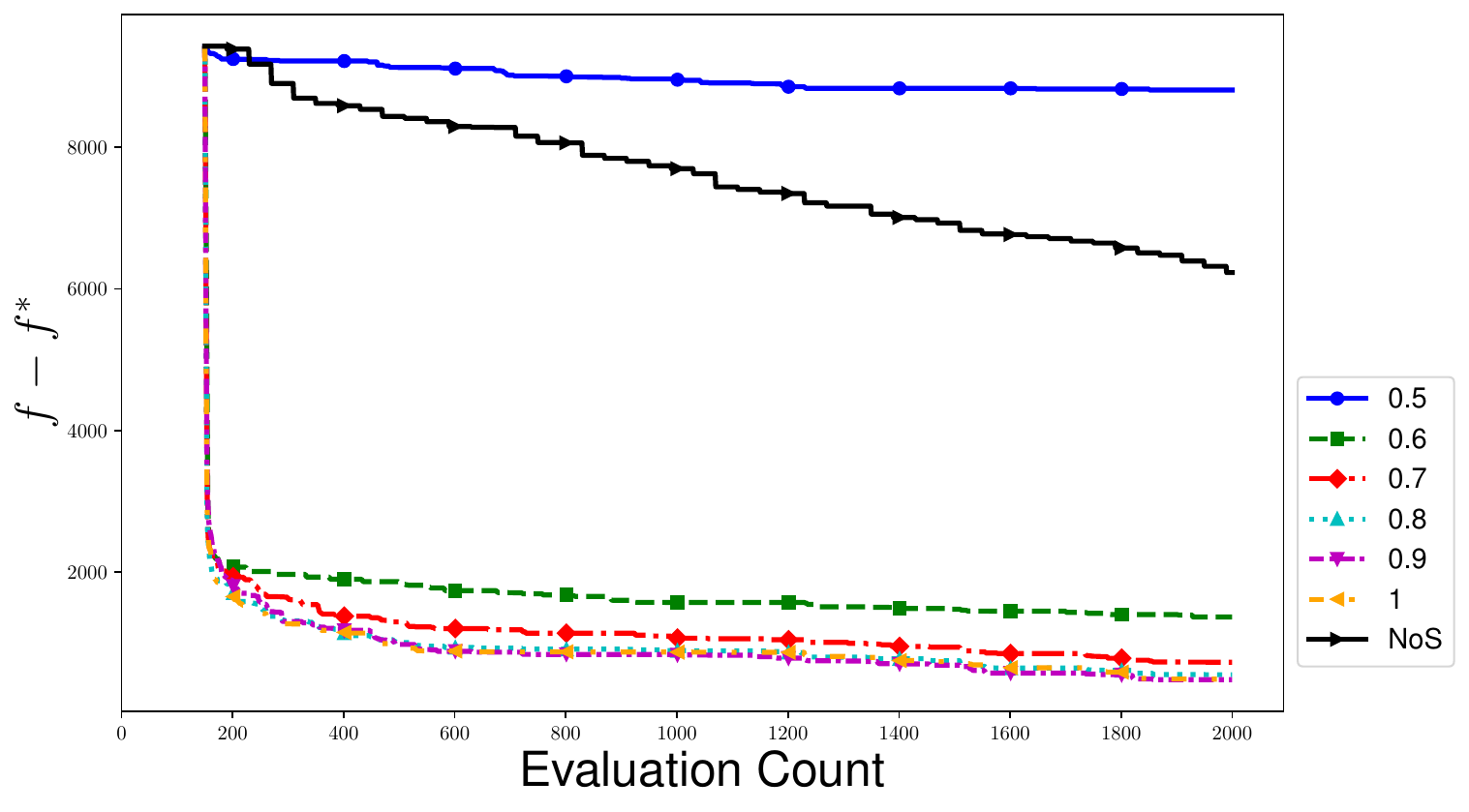}\label{fig:gbrbf_f4_d30}}
 \\
 \subfloat[$f8$]{\includegraphics[bb = 0 0 720 432, width=0.32\linewidth]{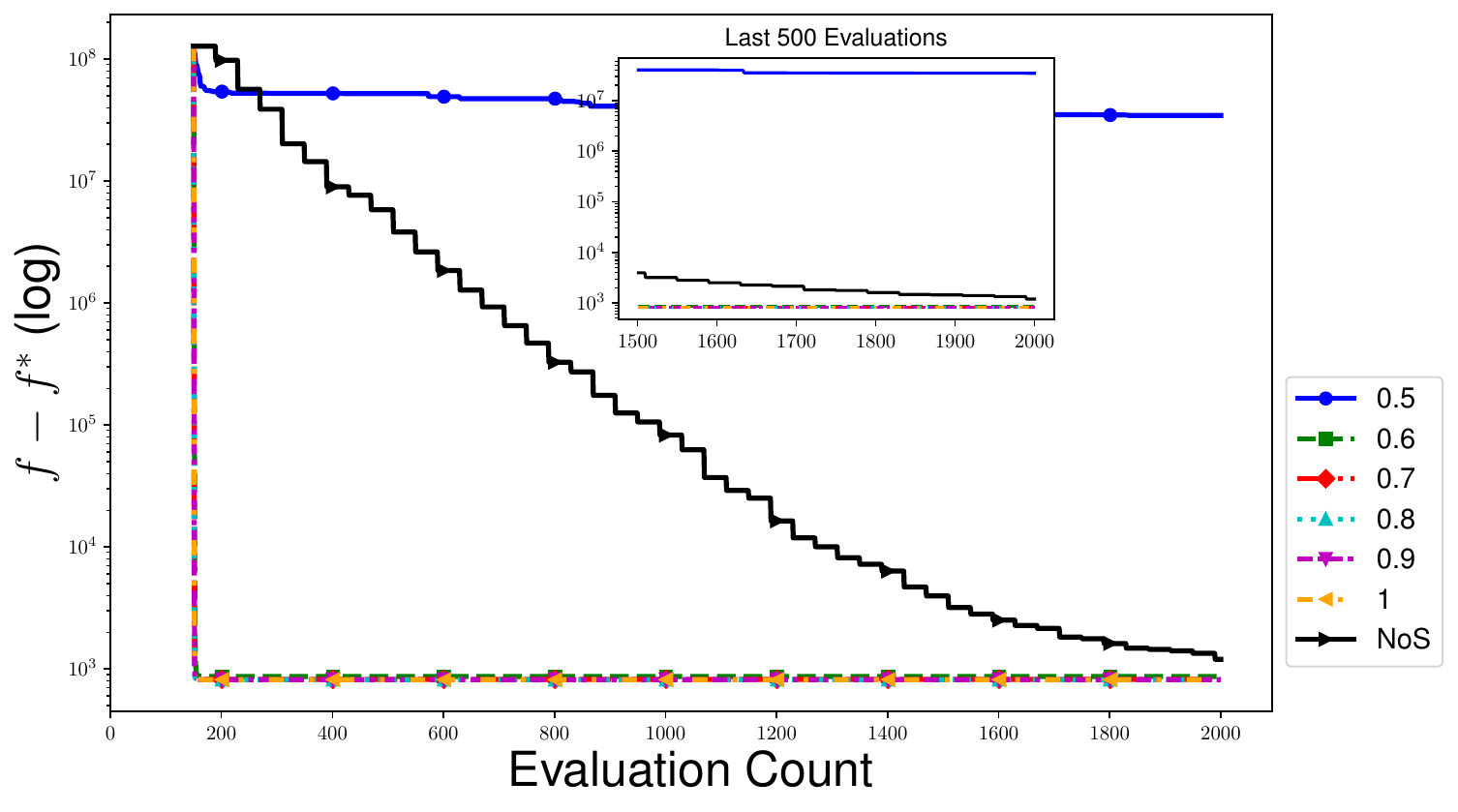}\label{fig:gbrbf_f8_d30}}
 \hfill
 \subfloat[$f13$]{\includegraphics[bb = 0 0 720 432, width=0.32\linewidth]{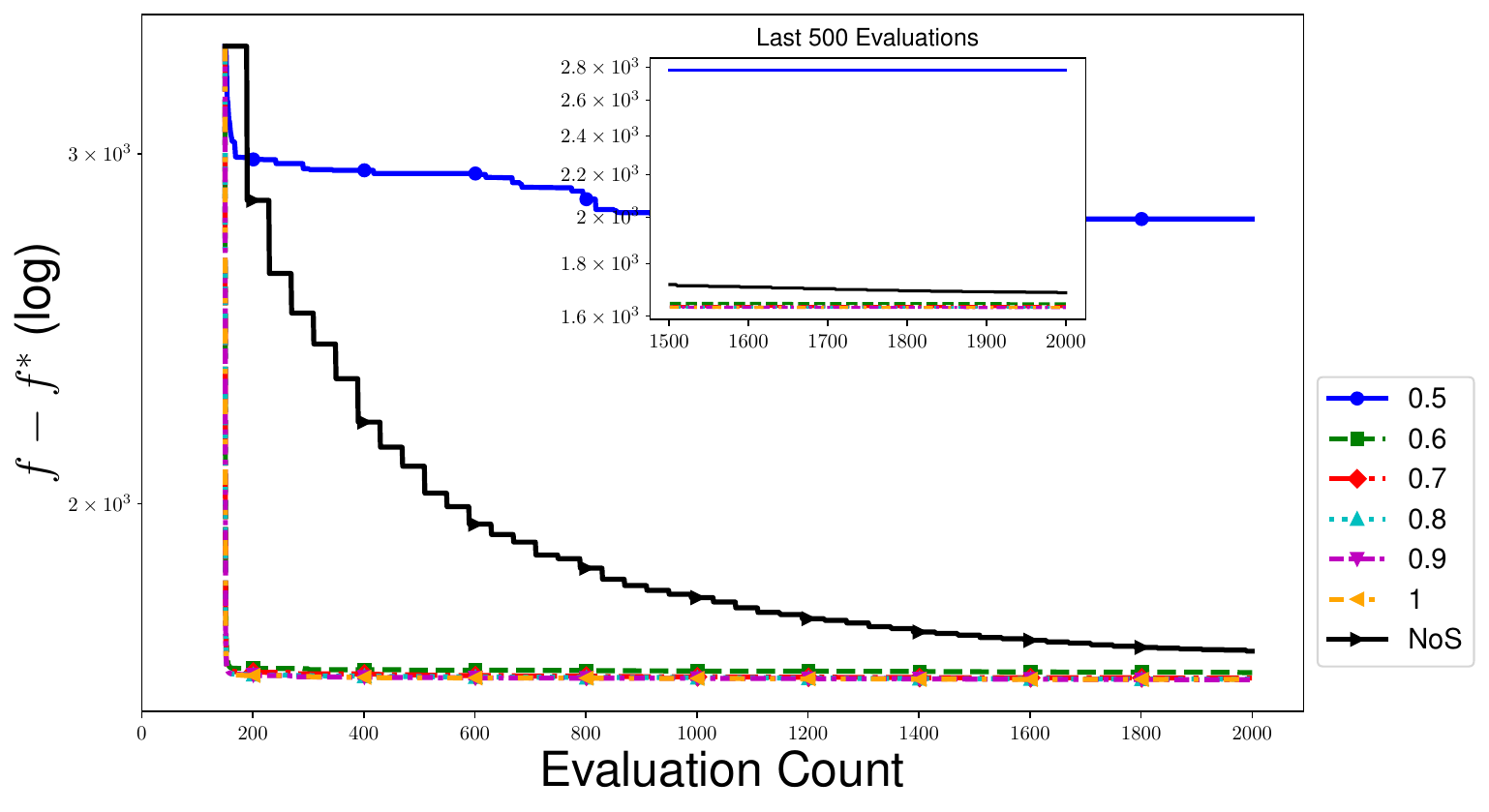}\label{fig:gbrbf_f13_d30}}
 \hfill
 \subfloat[$f15$]{\includegraphics[bb = 0 0 720 432, width=0.32\linewidth]{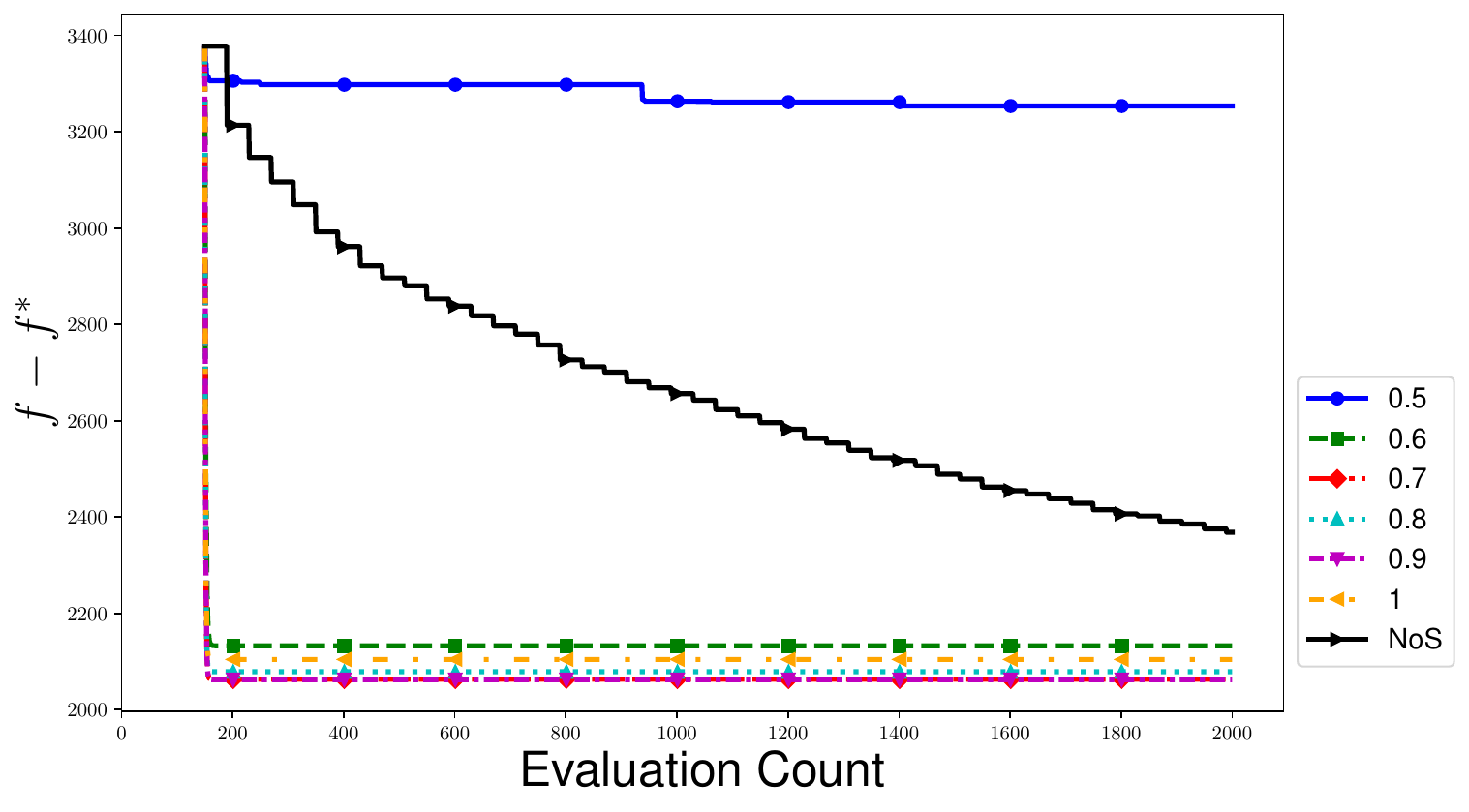}\label{fig:gbrbf_f15_d30}}
 \caption{Transition of the difference between the optimal objective function value and the value obtained by algorithms when solving 30D problems using GB}
 \label{fig:gbrbf_d30}
\end{figure*}

Figures \ref{fig:gbrbf_f1_d30} and \ref{fig:gbrbf_f2_d30} show the search results for unimodal problems. For $f1$, the smallest objective function value was obtained with an accuracy of 1.0. Next, an accuracy of 0.8 had good search performance. In addition, the search performance with an accuracy of 0.5 was worse than that of NoS.
However, for $f2$, the search performance was the best, with an accuracy of 0.8, and the second-best performance was achieved with an accuracy of 0.9. Furthermore, although it had an accuracy of 0.5, it could search for a better objective function value than the NoS at the maximum number of evaluations, and the NoS outperformed an accuracy of 0.5, up to approximately 700 evaluations.

Figures \ref{fig:gbrbf_f4_d30} and \ref{fig:gbrbf_f8_d30} show the results for multimodal problems. For $f4$, an accuracy of 0.9 achieved the smallest objective function value. In addition, the search performance with an accuracy of 0.5 was worse than that of NoS. For $f8$, an accuracy of 0.9 achieved the smallest objective function value. An accuracy of 0.5 outperformed NoS to about 300 evaluations, but after that, NoS exceeded the search performance with an accuracy of 0.5.

Figures \ref{fig:gbrbf_f13_d30} and \ref{fig:gbrbf_f15_d30} present the composite function results. For $f13$, an accuracy of 0.9 achieved the smallest objective function value, followed by an accuracy of 1.0. For $f15$, an accuracy of 0.9 achieved small values. For $f13$ and $f15$, an accuracy of 0.5 was better than NoS until about 200 evaluations, but after that, NoS outperformed an accuracy of 0.5.

\subsubsection{Correlation between Accuracy and Search Performance}
\begin{table*}[tb]
\centering
\caption{Kendall's rank correlation coefficient between the accuracy of the pseudo-surrogate model and the objective function value after 2,000 evaluations}
\begin{tabular}{cccccccc} 
\toprule
 & & $f1$ & $f2$ & $f4$ & $f8$ & $f13$ & $f15$ \\ 
\midrule
\multirow{2}{*}{PS} & 10D & 1.00 & 1.00 & 1.00 & 1.00 & 1.00 & 1.00 \\ \cmidrule{2-8}
 & 30D & 1.00 & 1.00 & 1.00& 1.00 & 1.00 & 1.00 \\ \midrule
\multirow{2}{*}{IB} & 10D & 0.47 & 0.33 & 0.60 & 0.33 & 0.93 & 0.73 \\ \cmidrule{2-8}
 & 30D & 0.87 & 0.60 & 0.87 & 0.60 & 0.87 & 0.33 \\ \midrule
\multirow{2}{*}{GB} & 10D & 0.73 & 0.87 & 0.89 & 0.65 & 0.97 & 1.00 \\ \cmidrule{2-8}
 & 30D & 0.87 & 0.60 & 0.87 & 0.87 & 0.87 & 0.47 \\
\bottomrule
\end{tabular}

\label{kendall's tau}
\end{table*}
Table~\ref{kendall's tau} shows Kendall's rank correlation coefficient between the accuracy of the pseudo-surrogate model and the objective function value after 2,000 evaluations. The coefficient ranged from $-1$ to $1$, where a value close to $-1$ indicated a negative correlation, which meant that a higher prediction accuracy reduced search performance in our experiment. However, a value close to one indicated a positive correlation, where higher accuracy resulted in better search performance. A coefficient of zero implied that there was no correlation. 

For PS, the results showed that higher accuracy led to better search performance across all functions in both 10 and 30 dimensions. For IB in 10 dimensions, the results on composite functions showed positive coefficients above 0.7, whereas those on unimodal and multimodal functions had coefficients below 0.6, indicating weak positive correlations. However, in 30 dimensions, although $f15$ showed a low coefficient of 0.33, the other cases demonstrated correlations greater than 0.6. For GB in 10 dimensions, although $f8$ showed a low value of 0.65, compared with other GB cases in 10 dimensions, positive correlations were observed overall. In 30 dimensions, while $f15$ showed a low coefficient of 0.47 compared with the other functions, the other functions exhibited positive correlations above 0.6.

\subsubsection{Summary}
From these results, we summarized this tendency as follows:
\begin{itemize}
 \item When using PS, the accuracy of surrogate models and search performance were fully correlated, and higher prediction accuracy consistently led to better performance.
 \item When using IB or GB, the accuracy and search performance showed a positive correlation. However, the highest accuracy did not always result in the best search performance, and lower accuracies might outperform higher accuracies.
\end{itemize}

From Algorithm~\ref{algorithm:original-PS-CS}, we could observe that in PS, offspring inferior to their parents were not selected for the next generation because if an inferior offspring was predicted to be superior to its parent, the actual evaluation revealed it to be inferior; therefore, it was rejected. Therefore, the low accuracy of PS did not contribute to increased diversity.

Conversely, from Algorithms~\ref{algorithm:original-IB-AFM} and \ref{algorithm:original-GB}, individuals with inferior evaluation values might be carried over to the next generation during sorting in IB and GB. This was because in Algorithm~\ref{algorithm:bubble_sort}, when one of the two individuals had not been evaluated with an actual evaluation function, it could be rearranged in the wrong order based on the surrogate accuracy. Owing to this characteristic, low accuracy might lead to increased diversity, potentially resulting in better search performance compared to higher accuracies.

\subsection{Sensitivity of Model Management Strategies to Prediction Accuracy}\label{sec:sen}
This section analyzed the sensitivity of the model management strategies to the prediction accuracy. Specifically, we confirmed a significant difference between the different prediction accuracies for each strategy by using Tukey's HSD test. 

\begin{figure*}[tb] 
 \centering
 \subfloat[$f1$]{\includegraphics[bb = 0 0 720 576, width=0.32\linewidth]{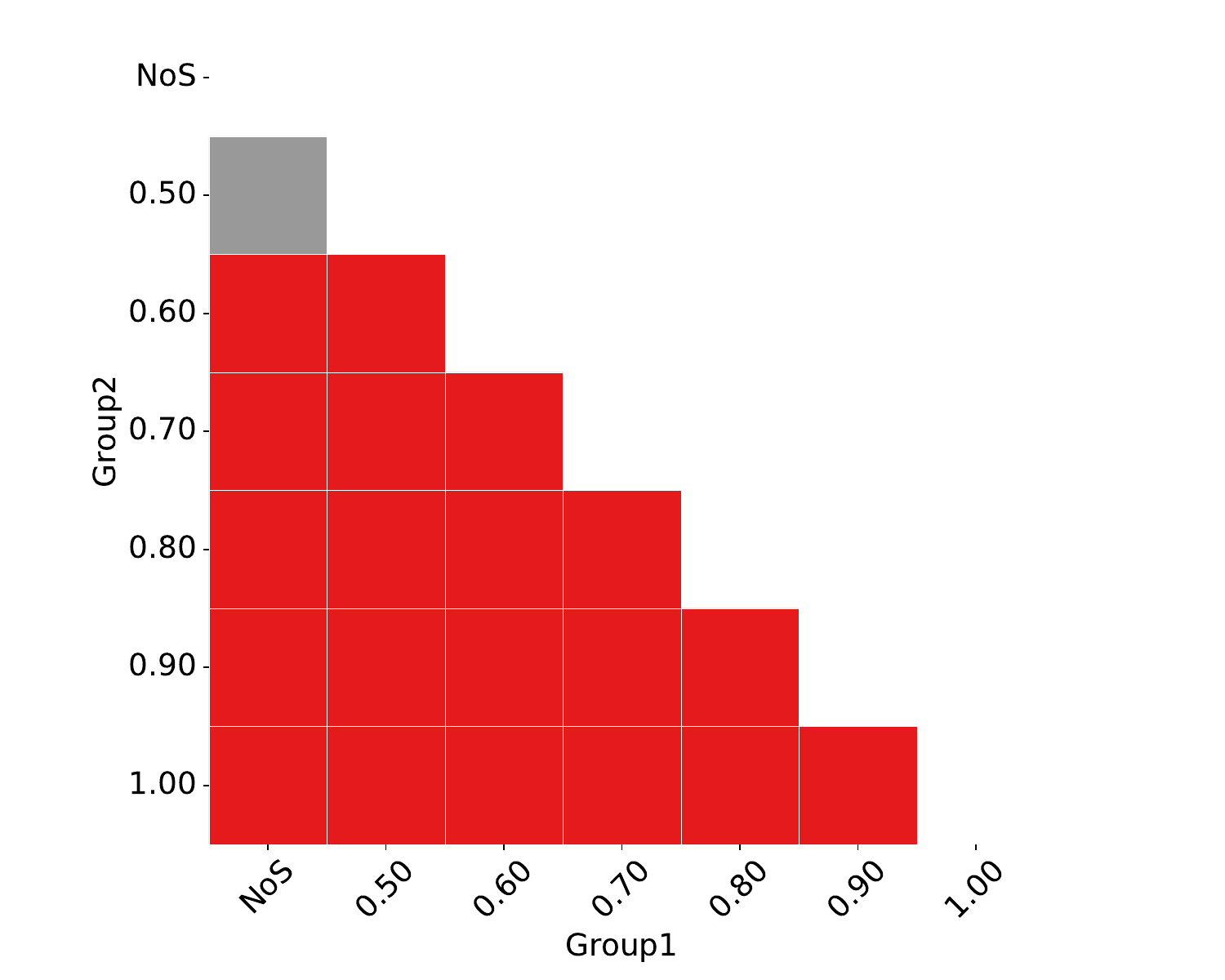}\label{fig:heat_pssvc_f1_d30}}
 \hfill
 \subfloat[$f2$]{\includegraphics[bb = 0 0 720 576, width=0.32\linewidth]{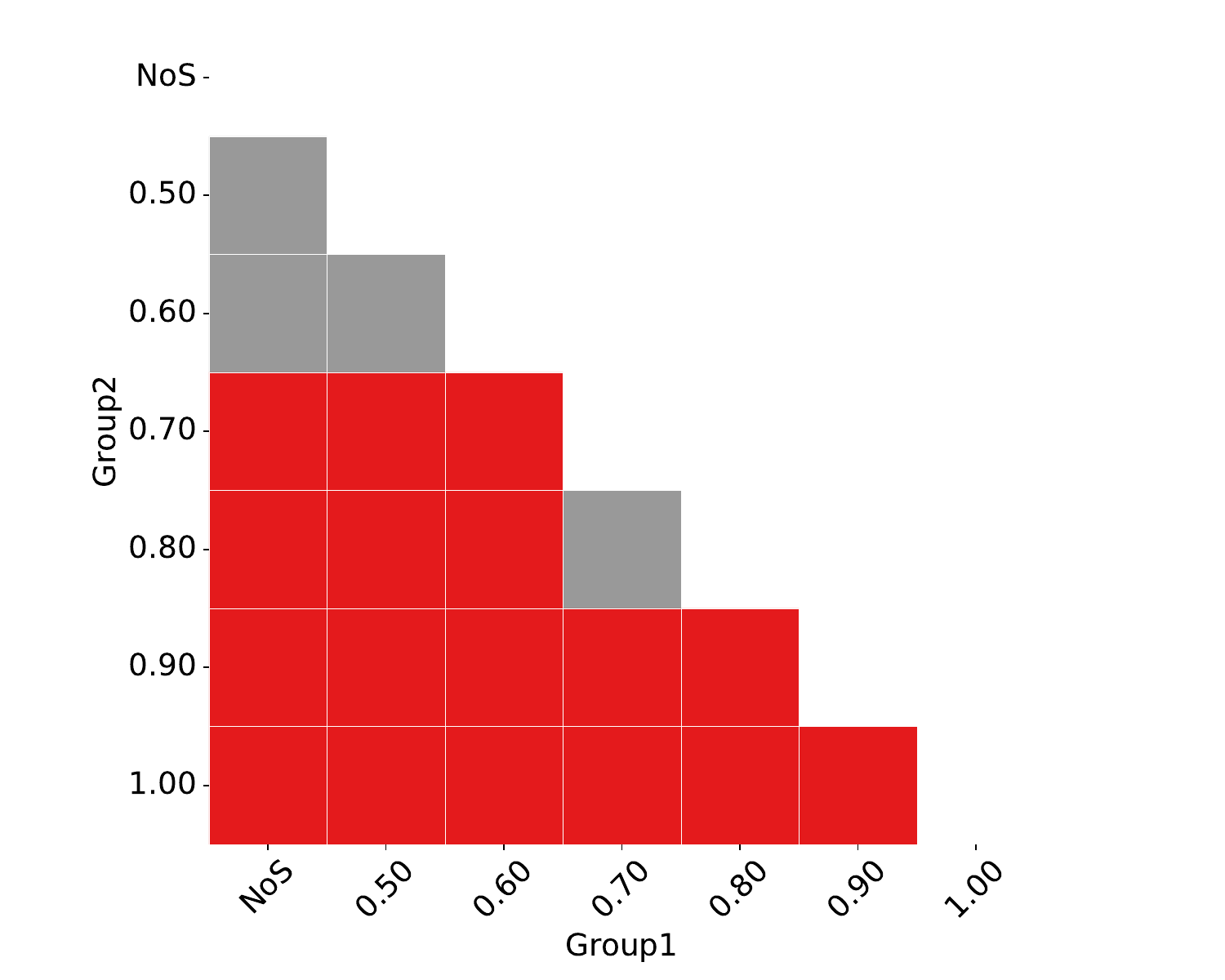}\label{fig:heat_pssvc_f2_d30}}
 \hfill
 \subfloat[$f4$]{\includegraphics[bb = 0 0 720 576, width=0.32\linewidth]{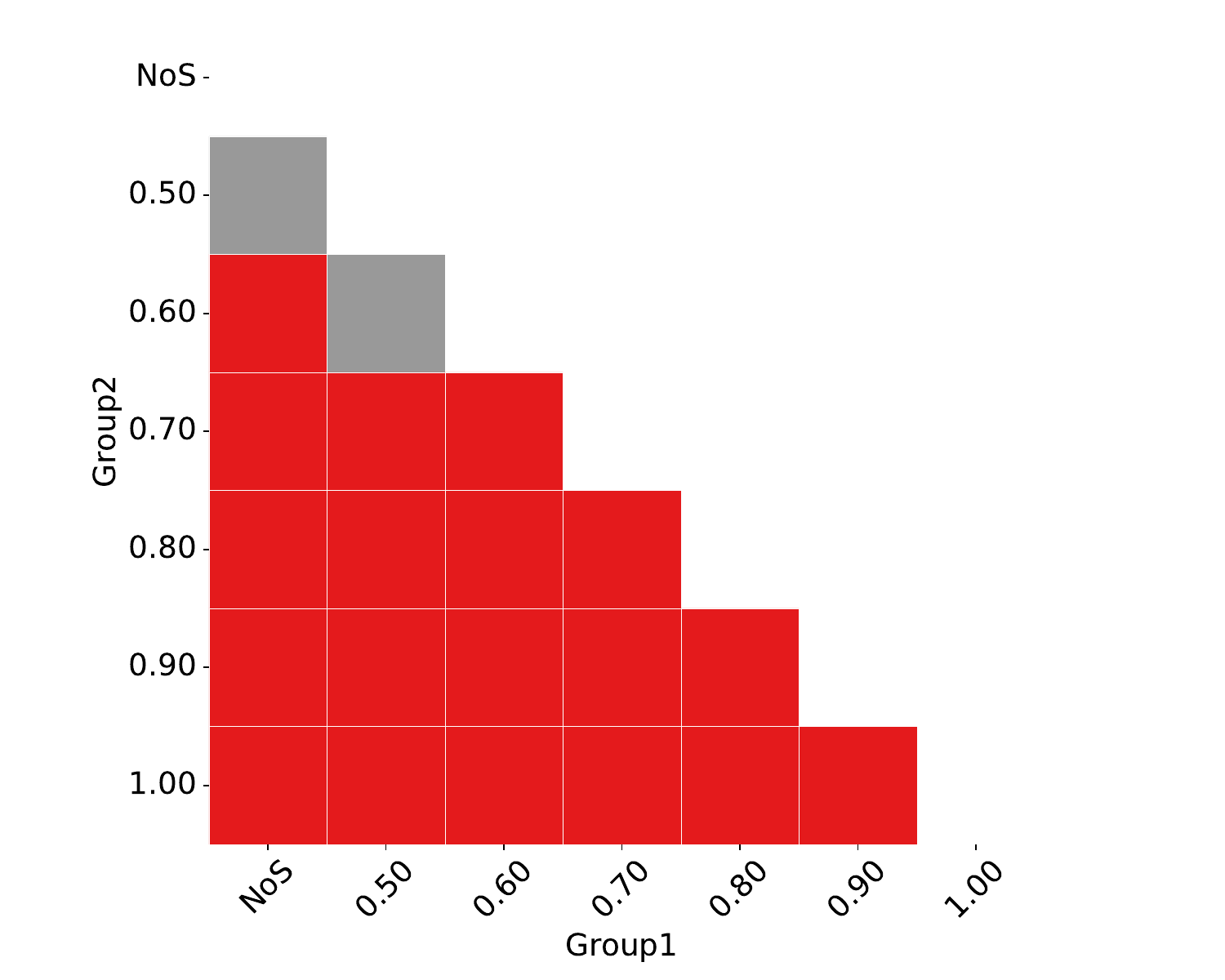}\label{fig:heat_pssvc_f4_d30}}
 \\
 \subfloat[$f8$]{\includegraphics[bb = 0 0 720 576, width=0.32\linewidth]{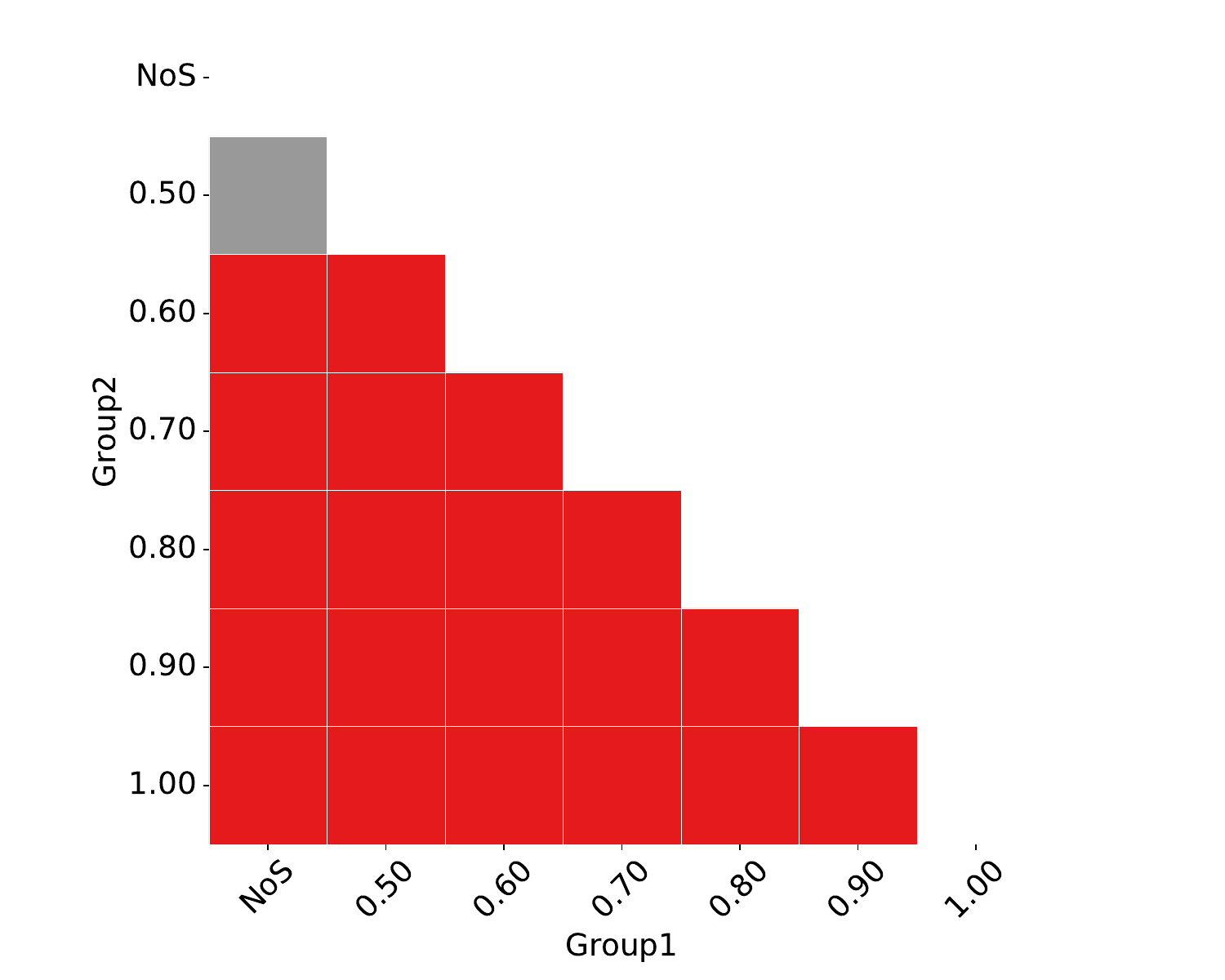}\label{fig:heat_pssvc_f8_d30}}
 \hfill
 \subfloat[$f13$]{\includegraphics[bb = 0 0 720 576, width=0.32\linewidth]{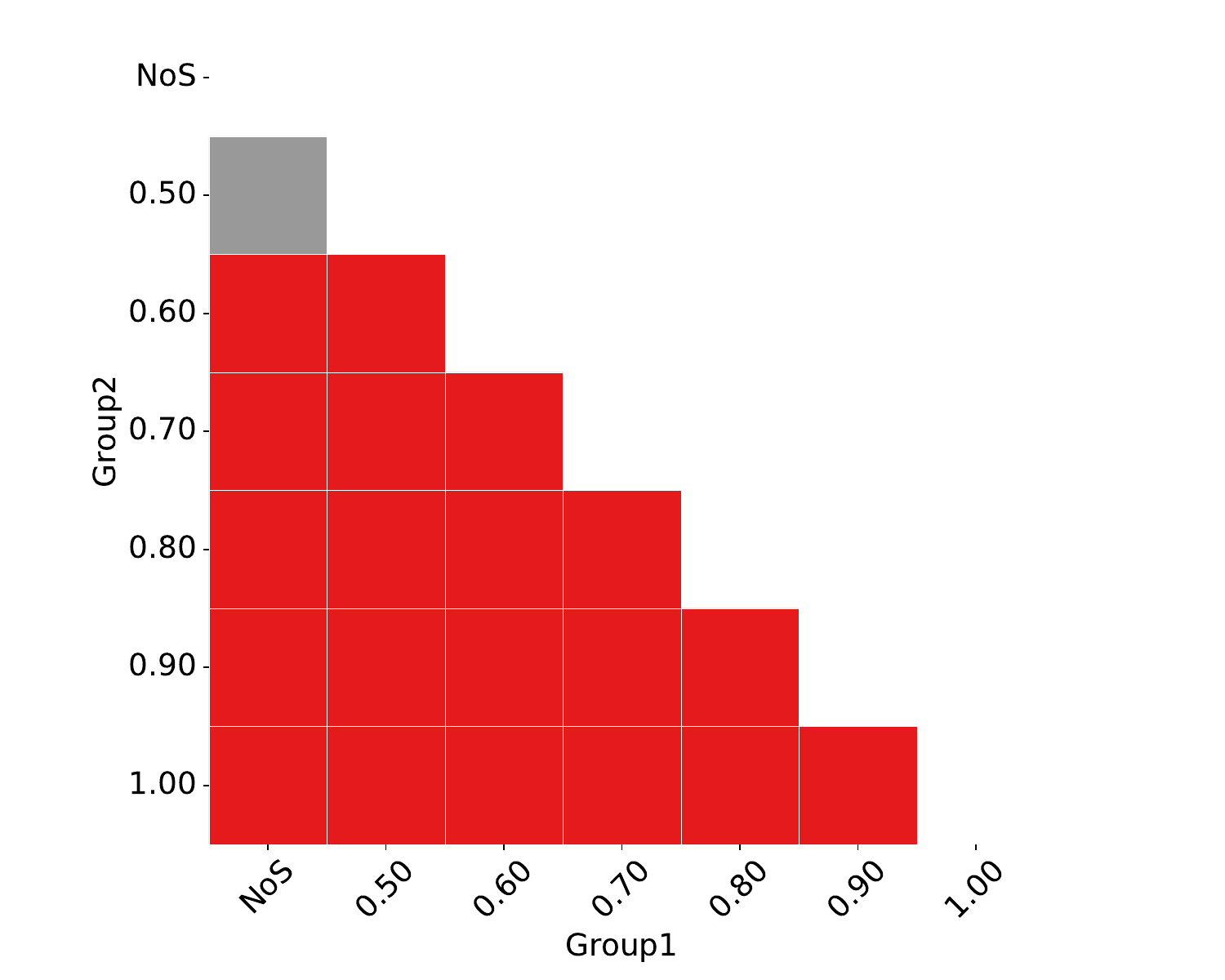}\label{fig:heat_pssvc_f13_d30}}
 \hfill
 \subfloat[$f15$]{\includegraphics[bb = 0 0 720 576, width=0.32\linewidth]{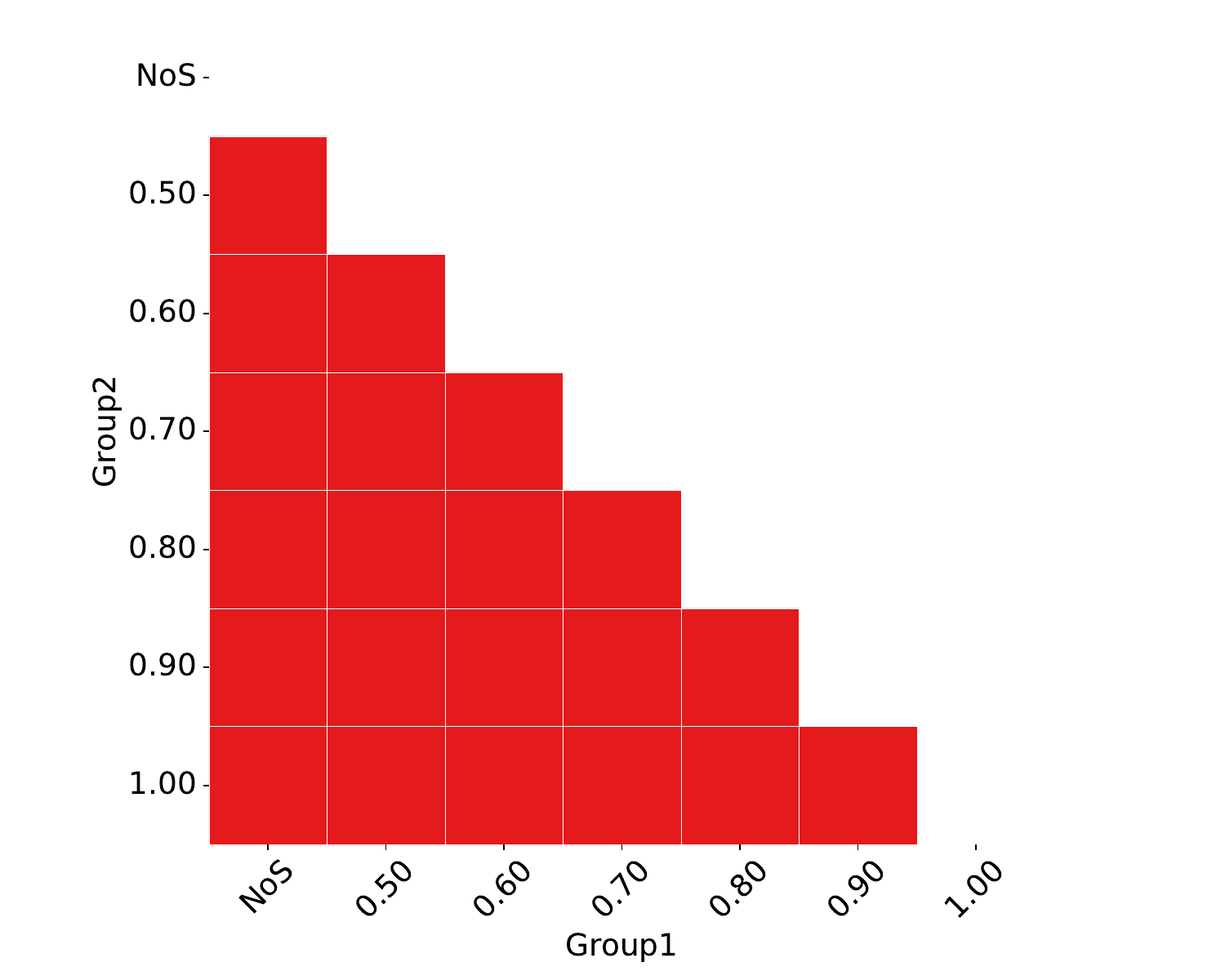}\label{fig:heat_pssvc_f15_d30}}
 \caption{Heatmap of Tukey's HSD test results between accuracies when searching 30D problems using PS}
 \label{fig:heat_pssvc_d30}
\end{figure*}

\begin{figure*}[tb] 
 \centering
 \subfloat[$f1$]{\includegraphics[bb = 0 0 720 576, width=0.32\linewidth]{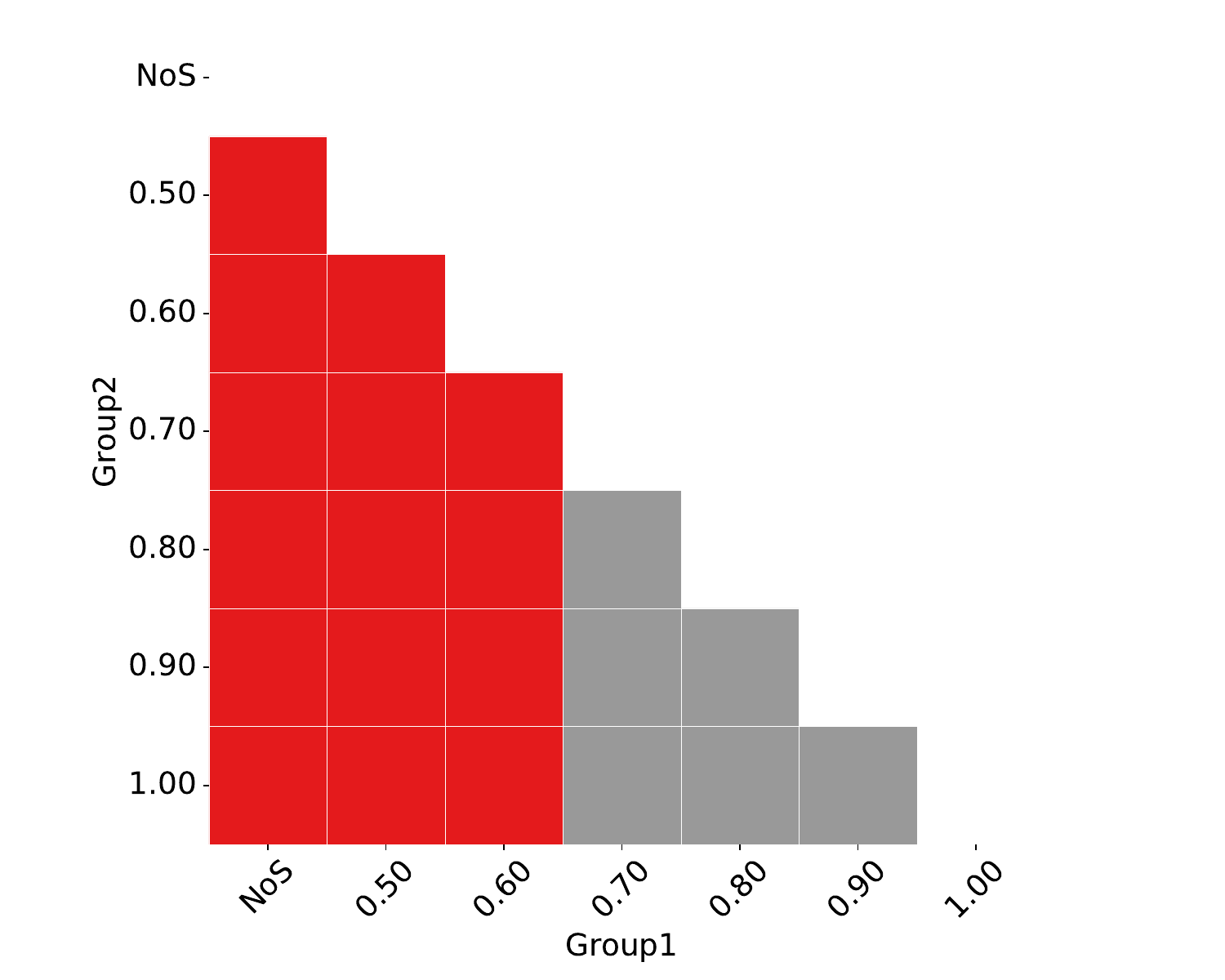}\label{fig:heat_ibafs_f1_d30}}
 \hfill
 \subfloat[$f2$]{\includegraphics[bb = 0 0 720 576, width=0.32\linewidth]{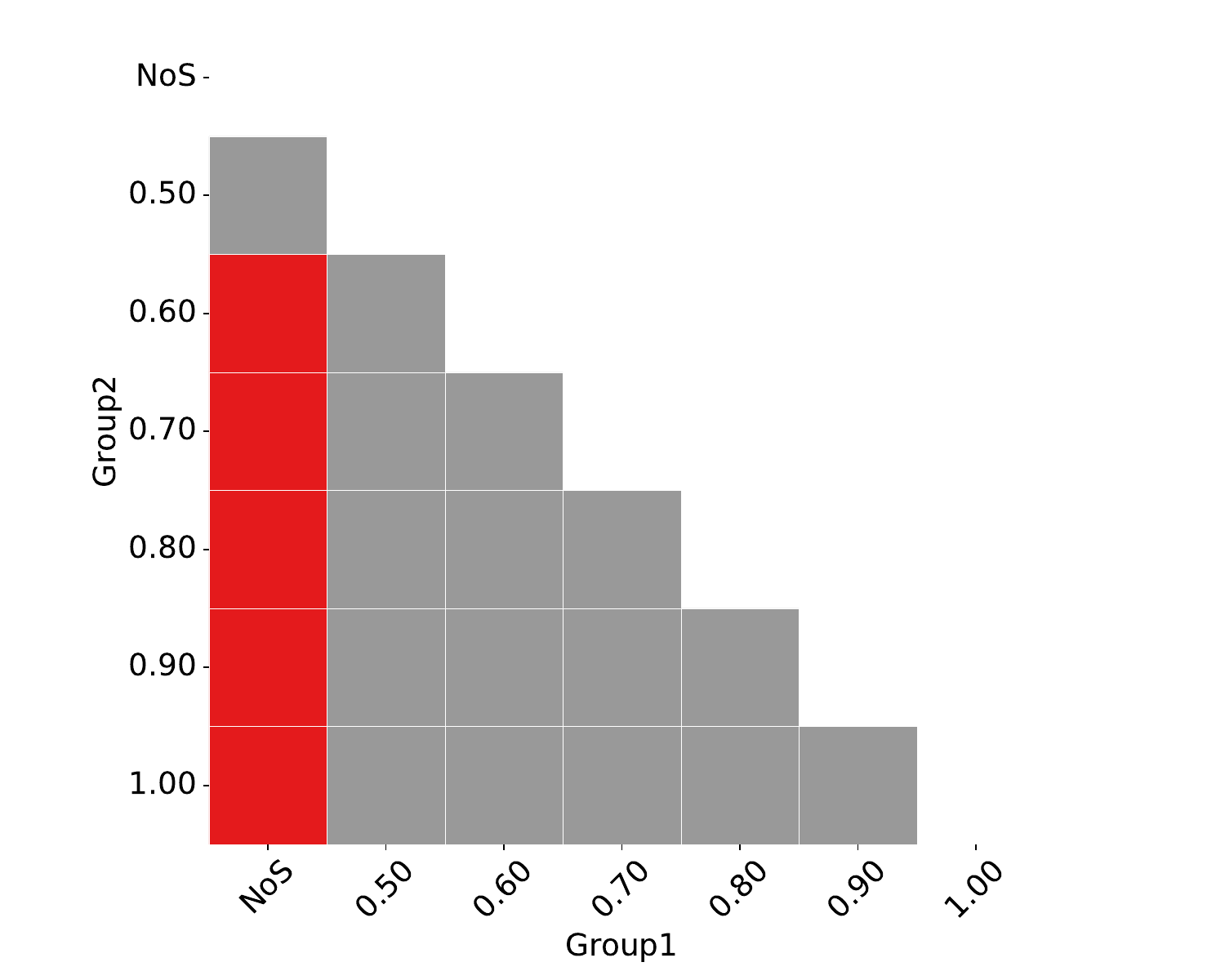}\label{fig:heat_ibafs_f2_d30}}
 \hfill
 \subfloat[$f4$]{\includegraphics[bb = 0 0 720 576, width=0.32\linewidth]{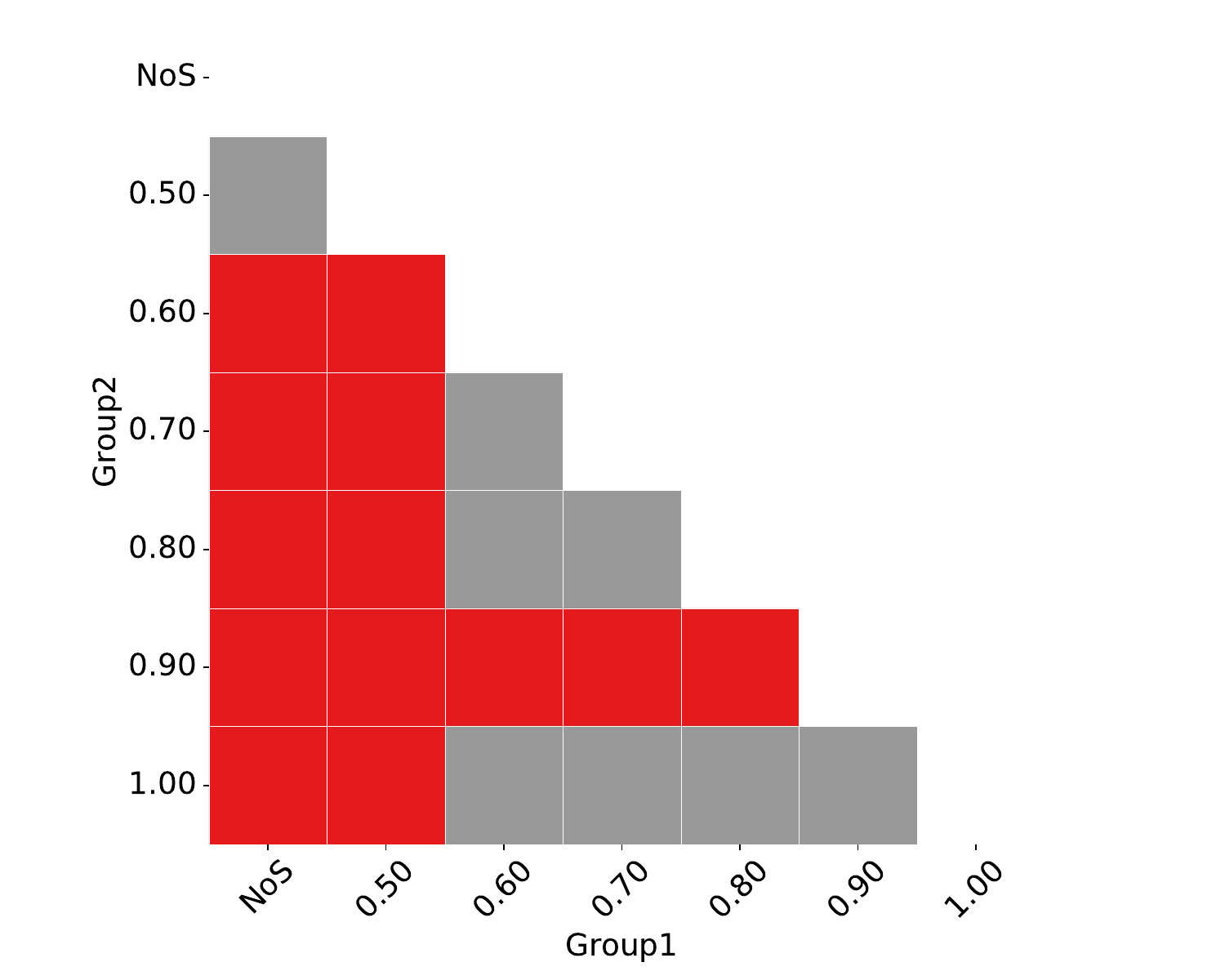}\label{fig:heat_ibafs_f4_d30}}
 \\
 \subfloat[$f8$]{\includegraphics[bb = 0 0 720 576, width=0.32\linewidth]{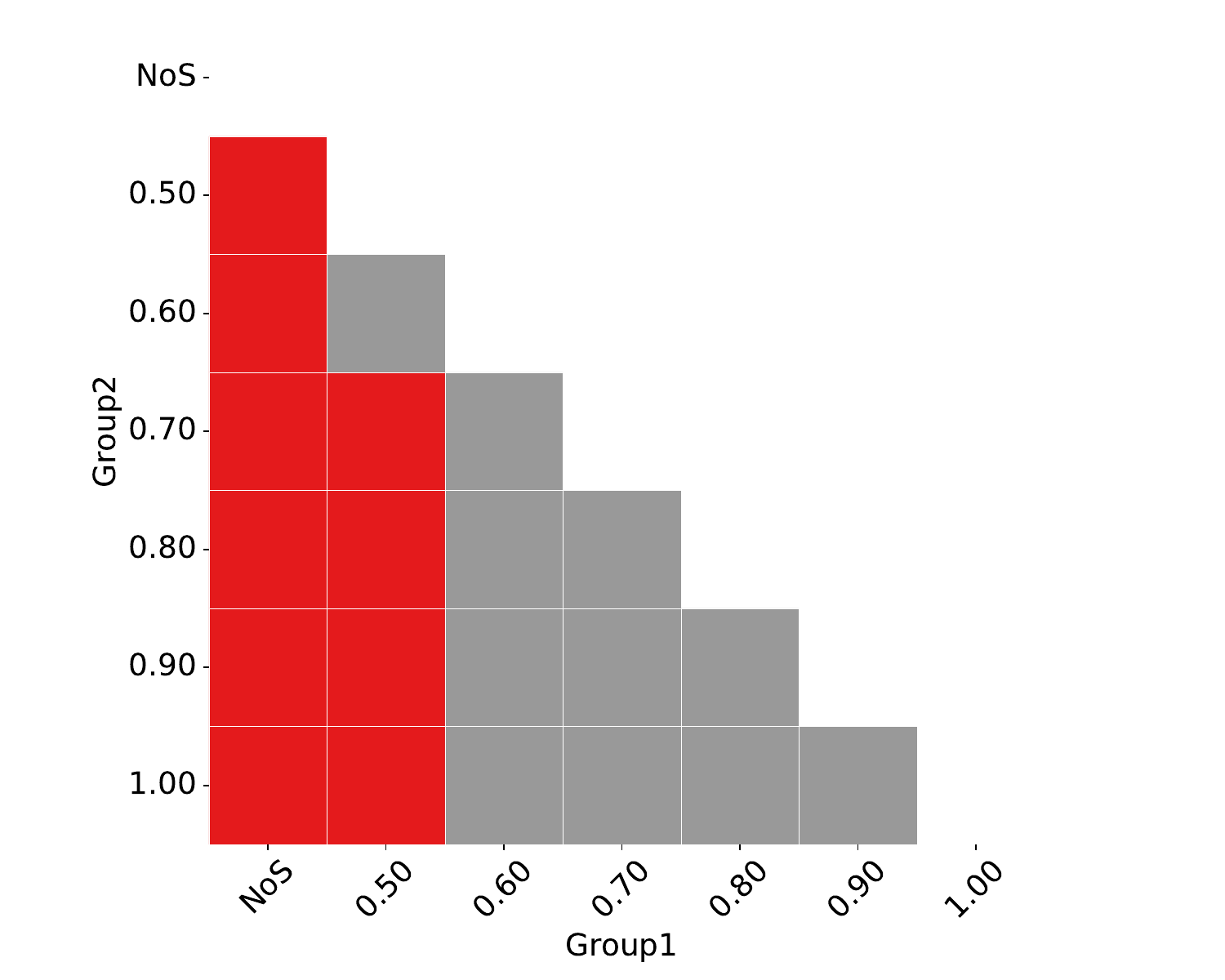}\label{fig:heat_ibafs_f8_d30}}
 \hfill
 \subfloat[$f13$]{\includegraphics[bb = 0 0 720 576, width=0.32\linewidth]{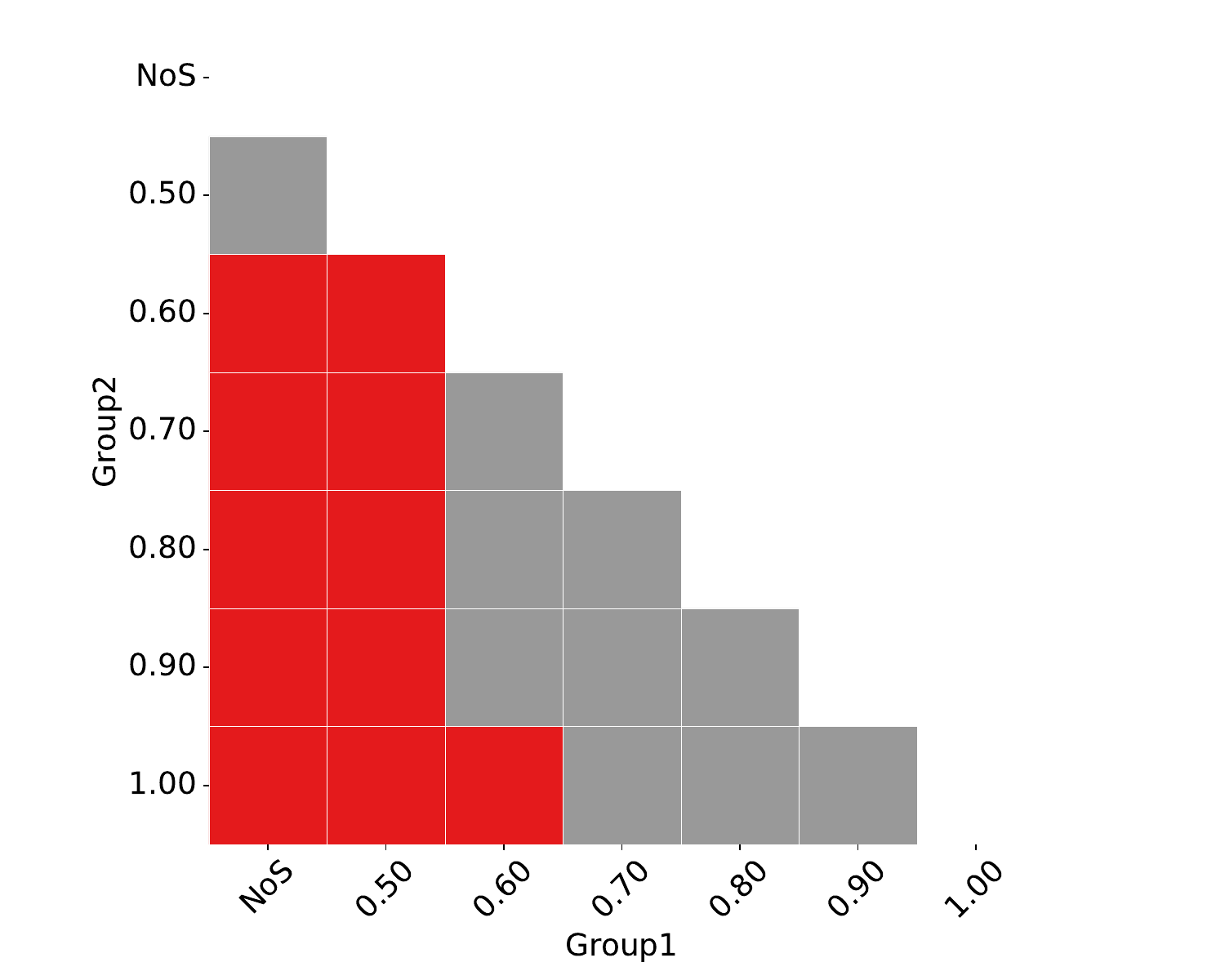}\label{fig:heat_ibafs_f13_d30}}
 \hfill
 \subfloat[$f15$]{\includegraphics[bb = 0 0 720 576, width=0.32\linewidth]{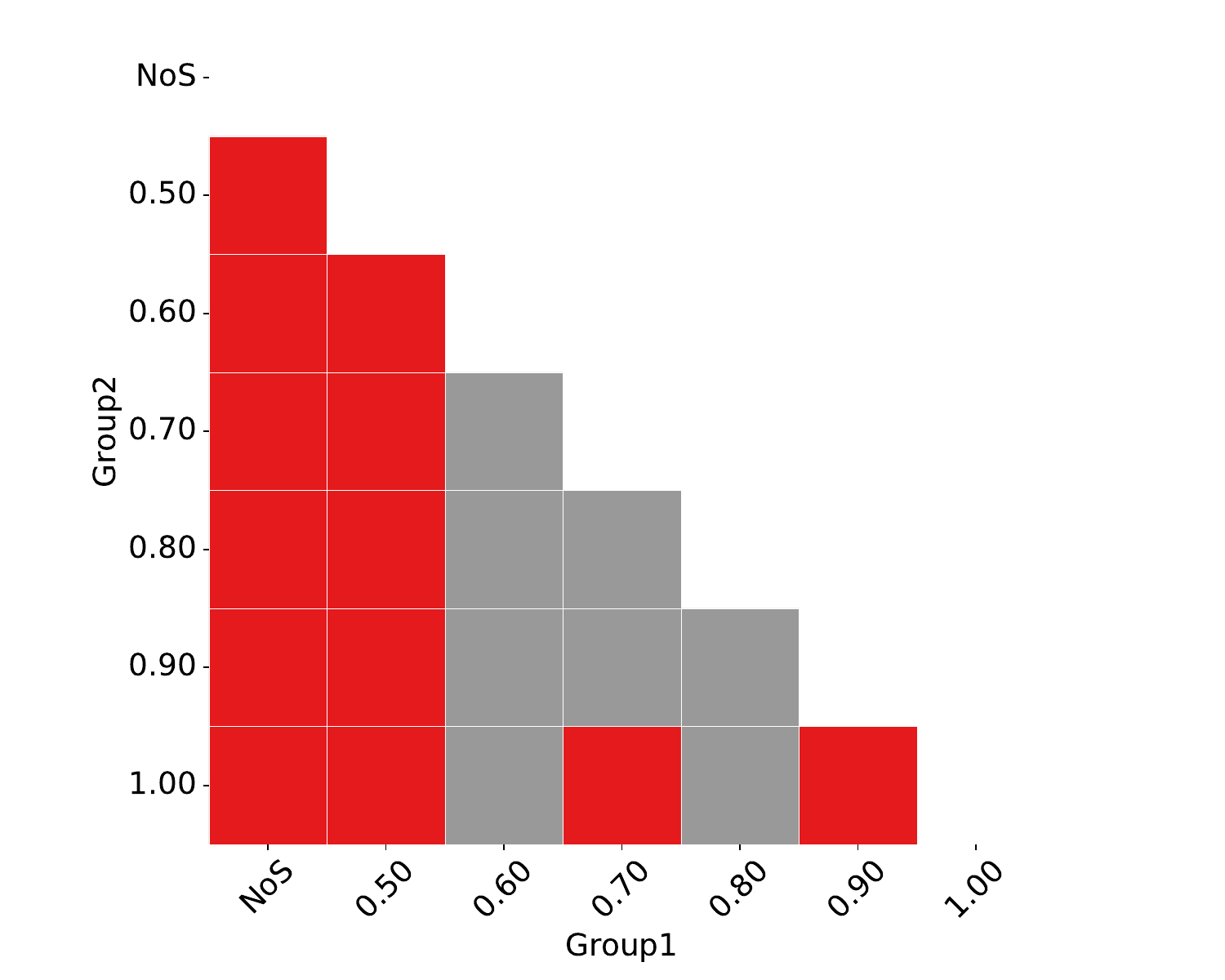}\label{fig:heat_ibafs_f15_d30}}
 \caption{Heatmap of Tukey's HSD test results between accuracies when searching 30D problems using IB}
 \label{fig:heat_ibafs_d30}
\end{figure*}

\begin{figure*}[tb] 
 \centering
 \subfloat[$f1$]{\includegraphics[bb = 0 0 720 576, width=0.32\linewidth]{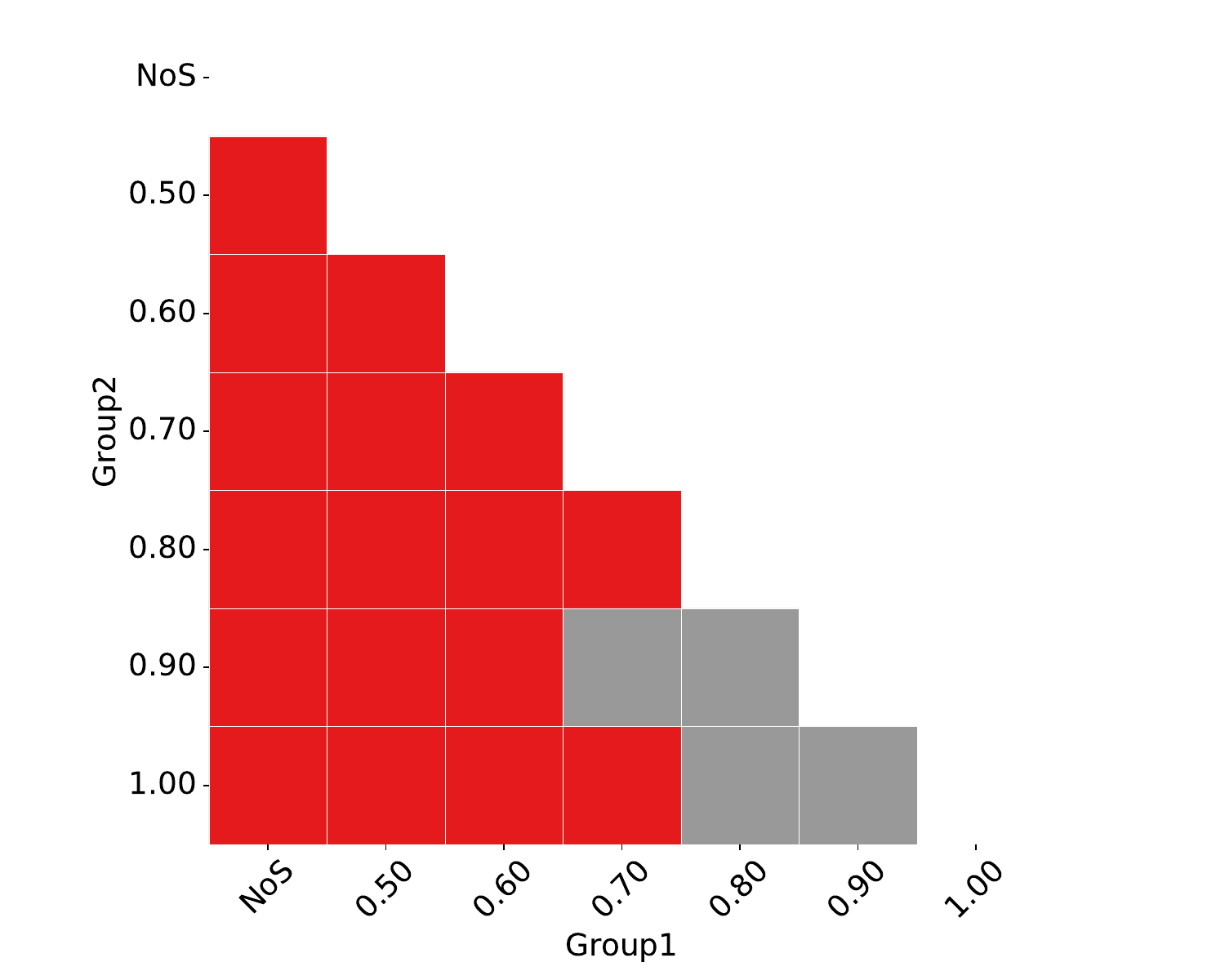}\label{fig:heat_gbafs_f1_d30}}
 \hfill
 \subfloat[$f2$]{\includegraphics[bb = 0 0 720 576, width=0.32\linewidth]{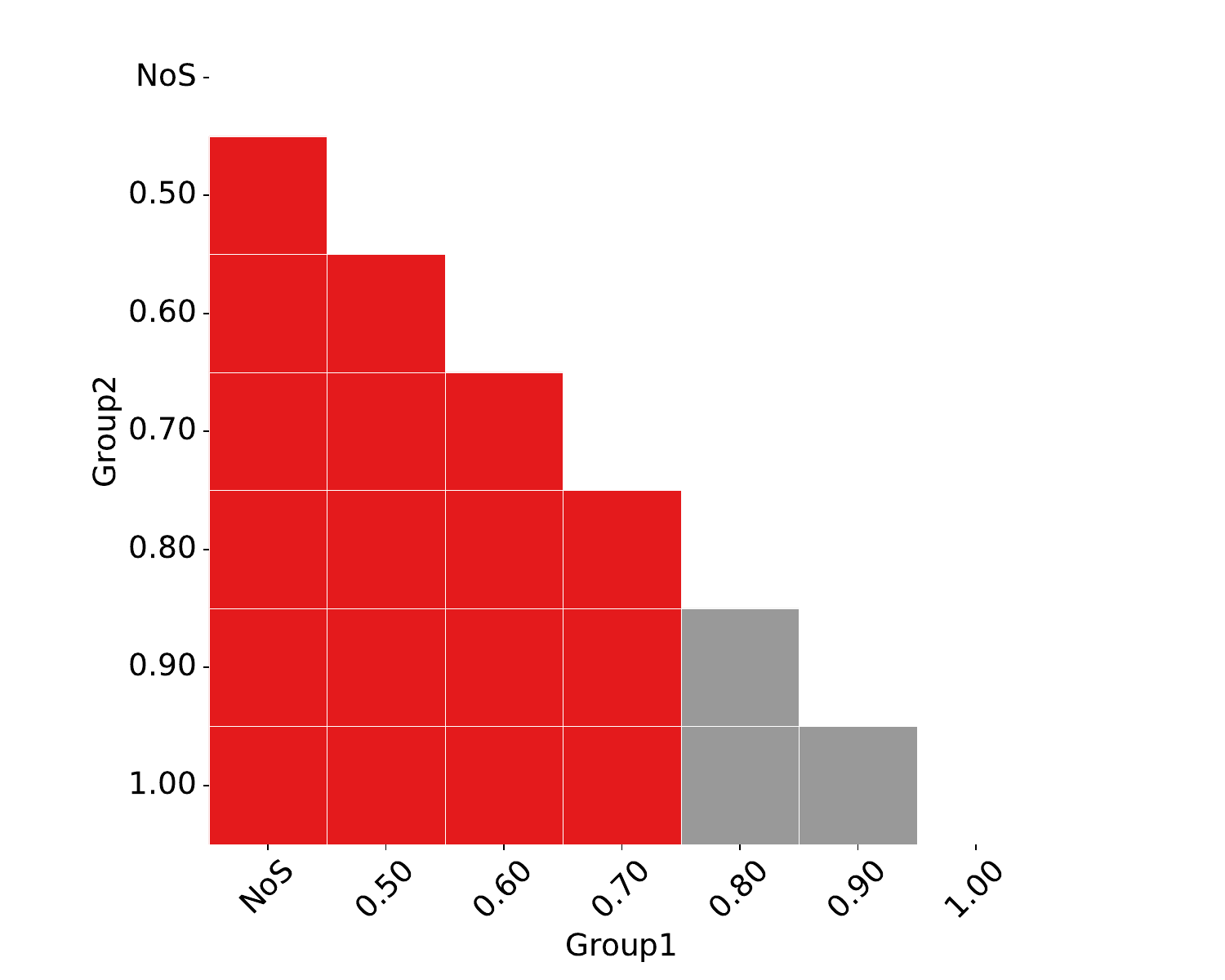}\label{fig:heat_gbafs_f2_d30}}
 \hfill
 \subfloat[$f4$]{\includegraphics[bb = 0 0 720 576, width=0.32\linewidth]{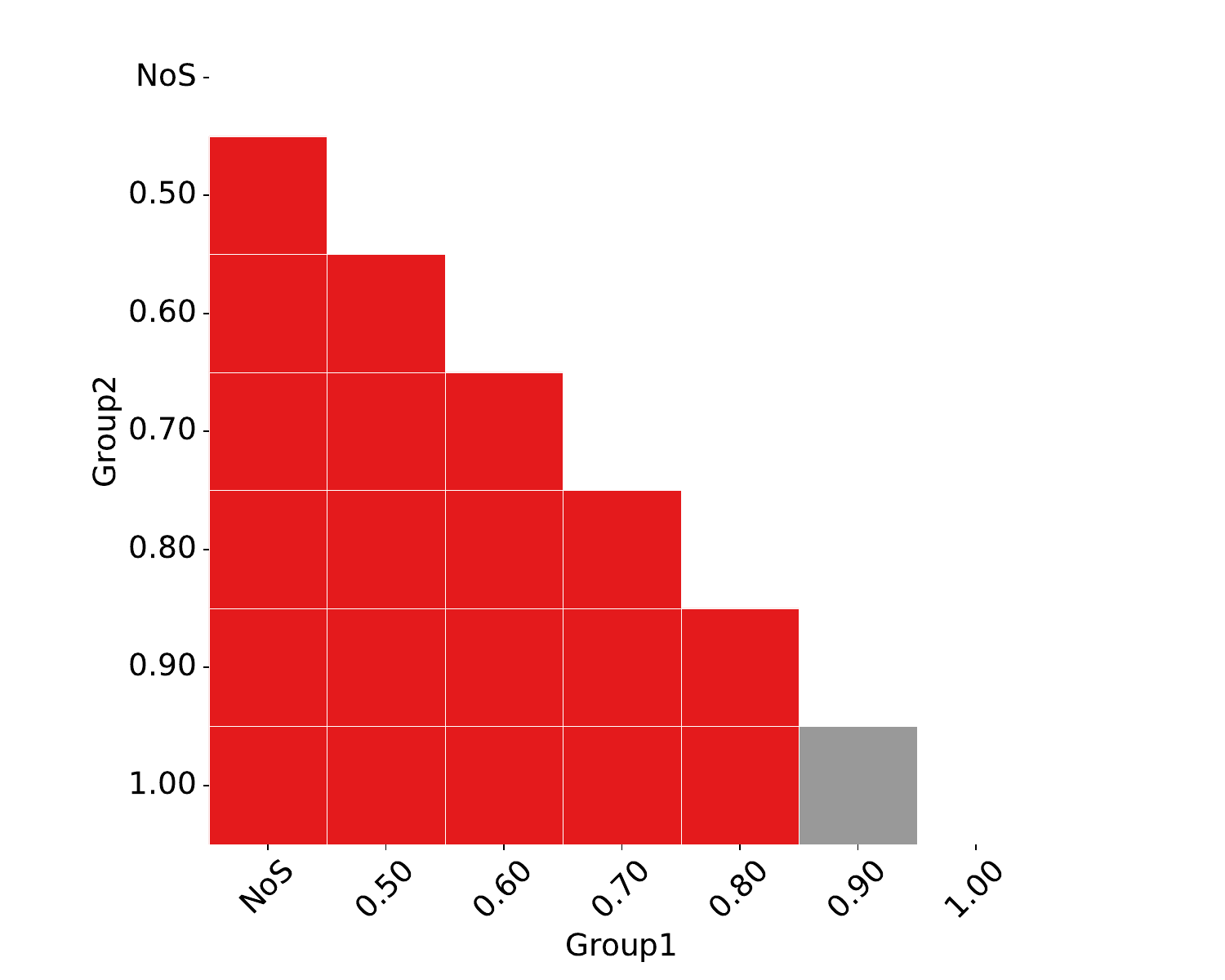}\label{fig:heat_gbafs_f4_d30}}
 \\
 \subfloat[$f8$]{\includegraphics[bb = 0 0 720 576, width=0.32\linewidth]{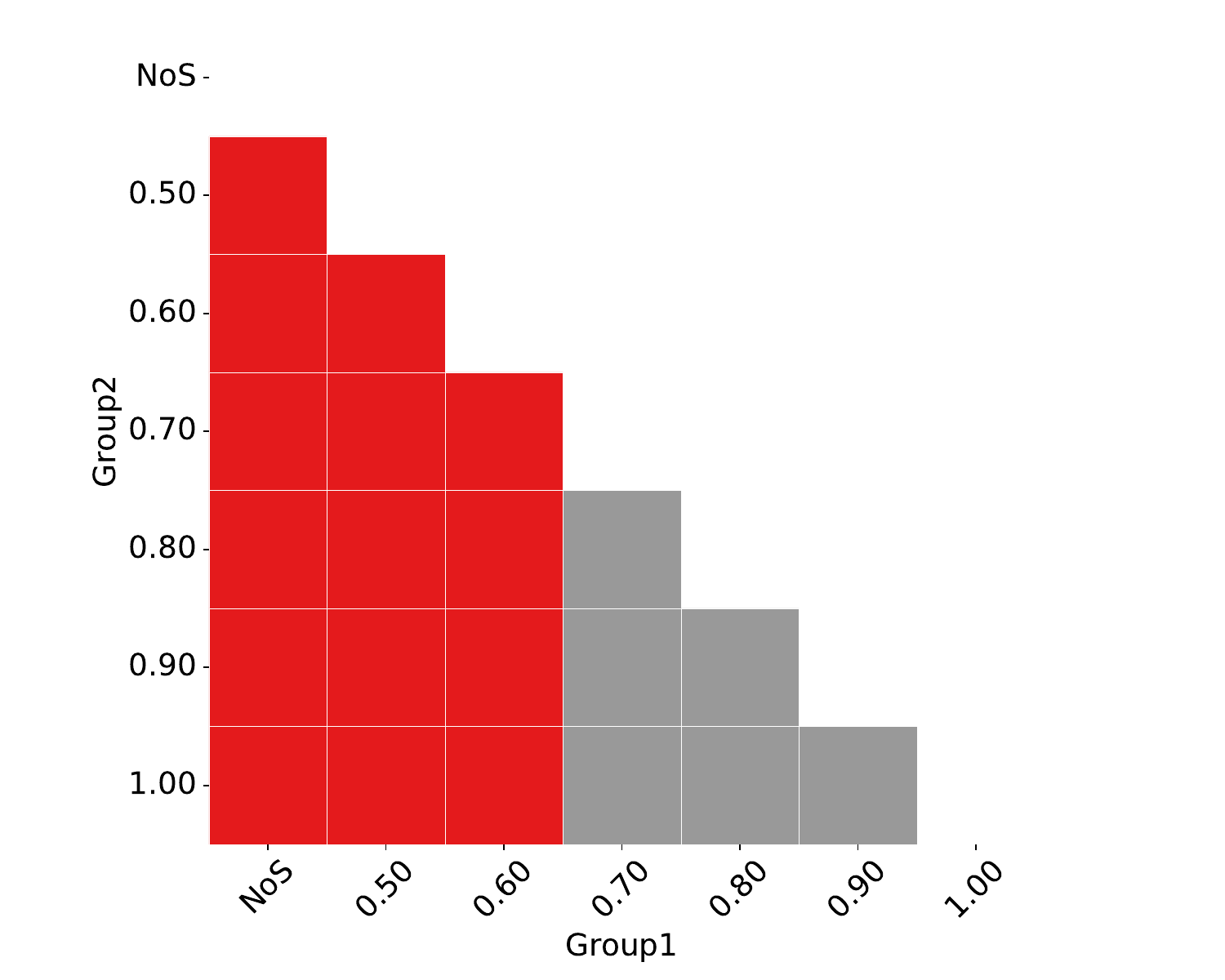}\label{fig:heat_gbafs_f8_d30}}
 \hfill
 \subfloat[$f13$]{\includegraphics[bb = 0 0 720 576, width=0.32\linewidth]{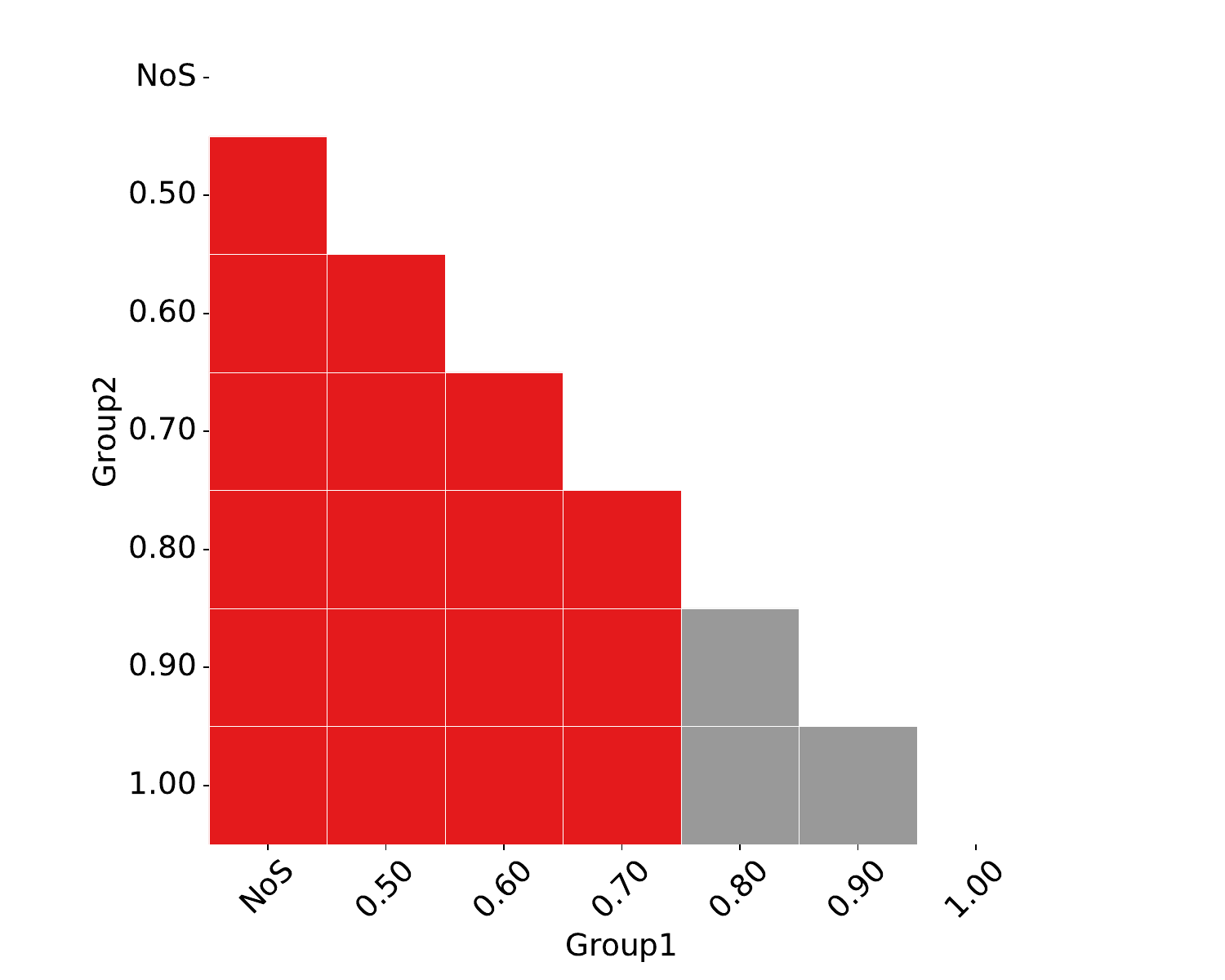}\label{fig:heat_gbafs_f13_d30}}
 \hfill
 \subfloat[$f15$]{\includegraphics[bb = 0 0 720 576, width=0.32\linewidth]{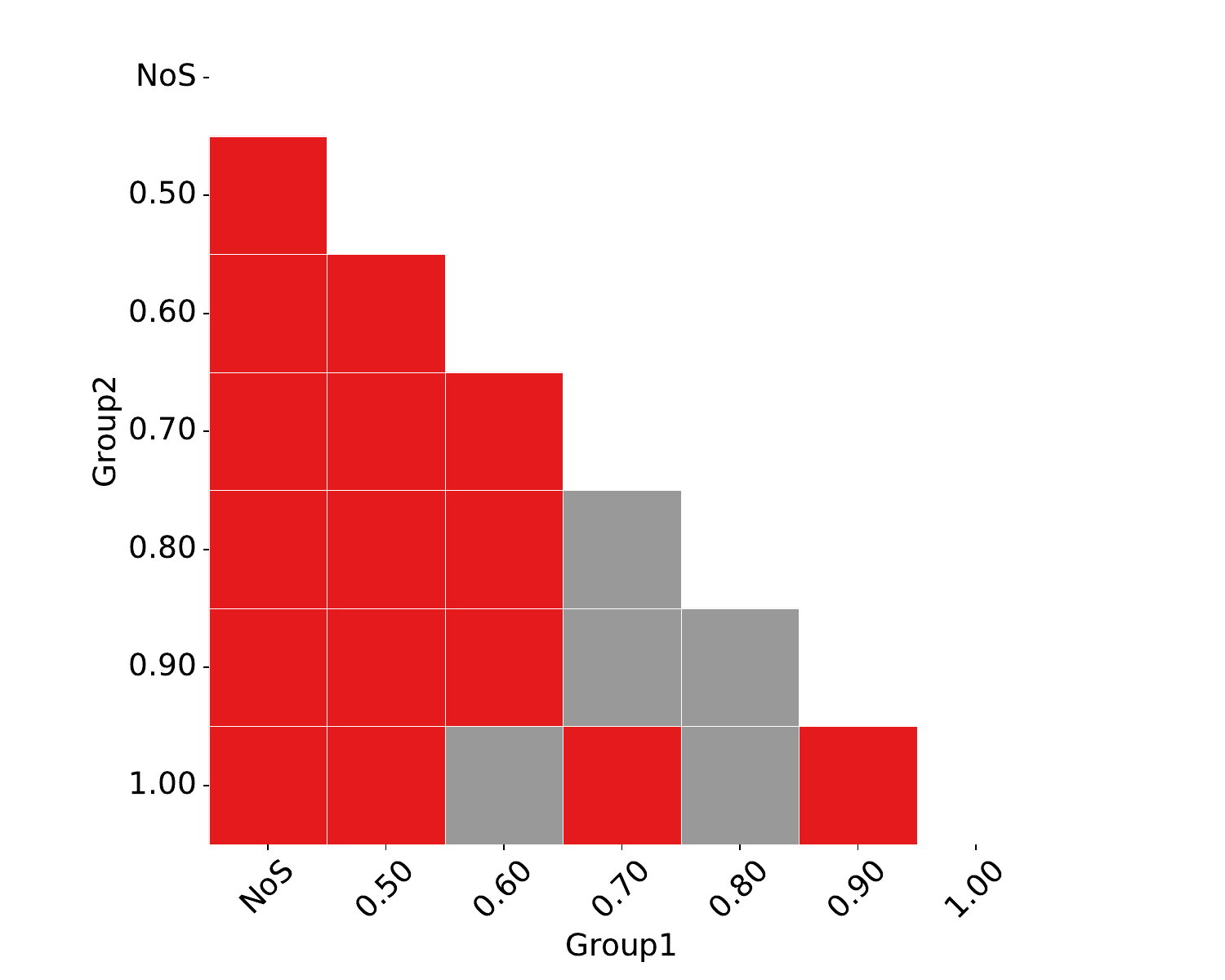}\label{fig:heat_gbafs_f15_d30}}
 \caption{Heatmap of Tukey's HSD test results between accuracies when searching 30D problems using GB}
 \label{fig:heat_gbafs_d30}
\end{figure*}

Figures \ref{fig:heat_pssvc_d30}--\ref{fig:heat_gbafs_d30} show the results of Tukey's HSD test between the objective function values at the maximum evaluations for each accuracy level across the 21 trials using PS, IB, and GB, respectively. Red indicated a significant difference between Groups 1 and 2, whereas blue indicated no significant difference.

The following subsections discuss the results of each model’s management strategy.
\subsubsection{PS}
For $f1$, $f8$, and $f13$, there was no significant difference between the accuracy of 0.5 and NoS, but significant differences could be confirmed between other accuracy pairs. For $f2$, no significant differences could be confirmed between the accuracies of 0.5 and 0.6, 0.5 and NoS, 0.6 and NoS, and 0.7 and 0.8. However, significant differences were confirmed when surrogates with higher accuracies were used. For $f4$, there were no significant differences between the accuracies of 0.5 and 0.6, and 0.5 and NoS, but significant differences were found in other pairs. For $f15$, all accuracy levels obtained significantly different results.

As a general trend, no significant difference existed between the results of the NoS and an accuracy level of 0.5, whereas the other combinations tended to show significant differences. This observation suggested that PS was an accuracy-sensitive strategy, implying that its search performance was influenced by the accuracy of the surrogate model used.

\subsubsection{IB}
For $f1$, there were no significant differences between the accuracies of 0.7 to 1.0, but there were significant differences between the other accuracy levels. For $f2$, significant differences only existed between the NoS and accuracies of 0.6 or higher, with no significant differences between other accuracy levels. For $f4$, there were no significant differences between the NoS and an accuracy of 0.5, between accuracies of 0.6 to 1.0 (excluding 0.9), and between accuracies of 0.9 and 1.0. For $f8$, there were no significant differences between the accuracies of 0.6 to 1.0 and between 0.5 and 0.6. For $f13$, there were no significant differences between the accuracies of 0.6 to 1.0 and between the accuracy of 0.5 and NoS. For $f15$, there were no significant differences between the accuracies of 0.6 to 0.9, 0.6, and 1.0, and 0.8 and 1.0.

These results indicated that the search performance of IB was robust, showing no significant difference when the surrogate accuracy was about 0.7 or higher. However, the performance declined when the accuracy decreased below 0.6.

\subsubsection{GB}
For $f1$, there were no significant differences between accuracies of 0.7 and 0.9, as well as between accuracies ranging from 0.8 to 1.0, but significant differences were found across other accuracies. For $f2$, accuracies between 0.8 and 1.0 showed no significant difference, while other accuracies showed significant differences. For $f4$, a significant difference was only observed between the accuracies of 0.9 and 1.0. For $f8$, no significant differences were found between the accuracies of 0.7 and 1.0. For $f13$, accuracies of 0.8 to 1.0 exhibited no significant differences. Finally, for $f15$, no significant differences were found between the accuracies of 0.6, 1.0, 0.7, 0.9, and 0.8 and 1.0.

Overall, there was a tendency for no significant difference to be observed among the search results when the accuracy was high. Specifically, the search performance of GB was robust when the accuracy was 0.8 or higher. In contrast, with an accuracy of 0.7 or lower, significant differences emerged, indicating that the search performance was more sensitive to low-accuracy surrogate models.

\subsection{Comparison of Different SAEA Strategies Across the Same Accuracy}\label{sec:com}
This section compares the search performance of the different model management strategies for the same accuracy. Specifically, we performed the Mann-Whitney U-test to compare the final results of PS, IB, and GB with the same accuracy.

\begin{figure*}[tb] 
 \centering
 \subfloat[PS vs IB on 10D]{\includegraphics[bb = 0 0 518 427, width=0.32\linewidth]{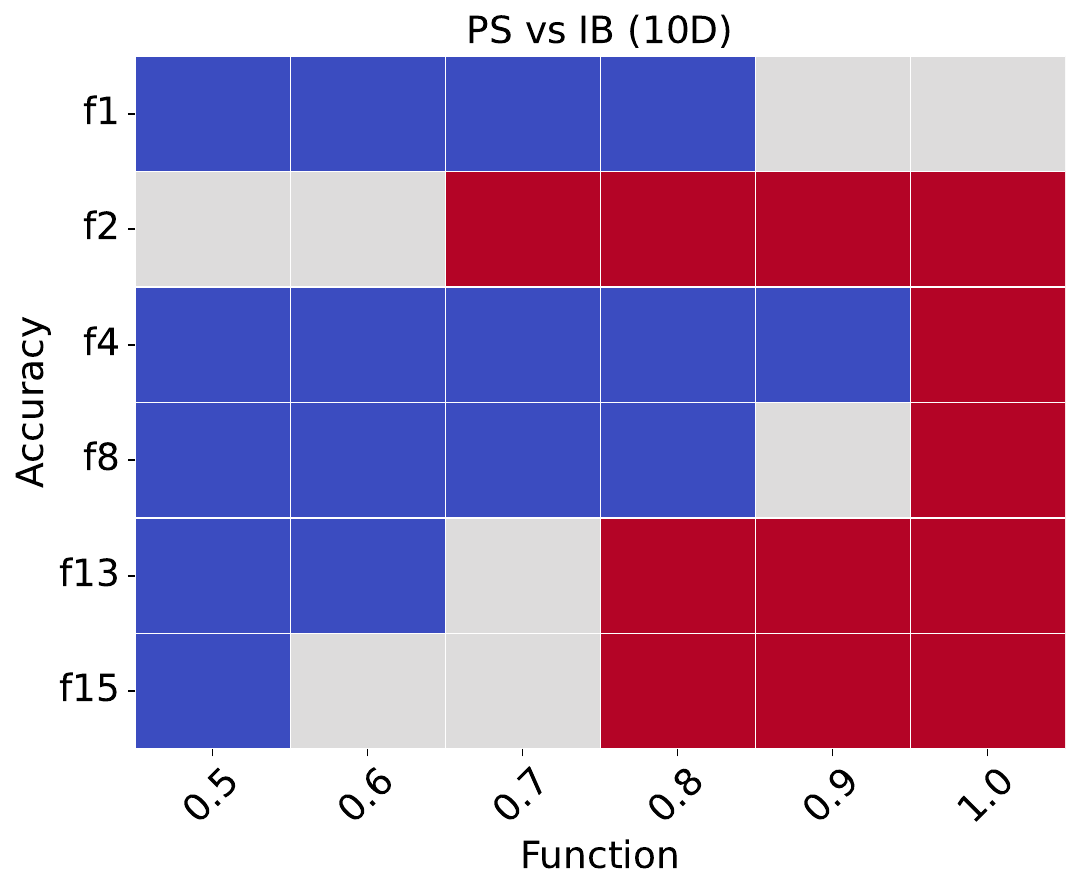}\label{fig:PS_vs_IB_10D}}
 \hfill
 \subfloat[PS vs GB on 10D]{\includegraphics[bb = 0 0 518 427, width=0.32\linewidth]{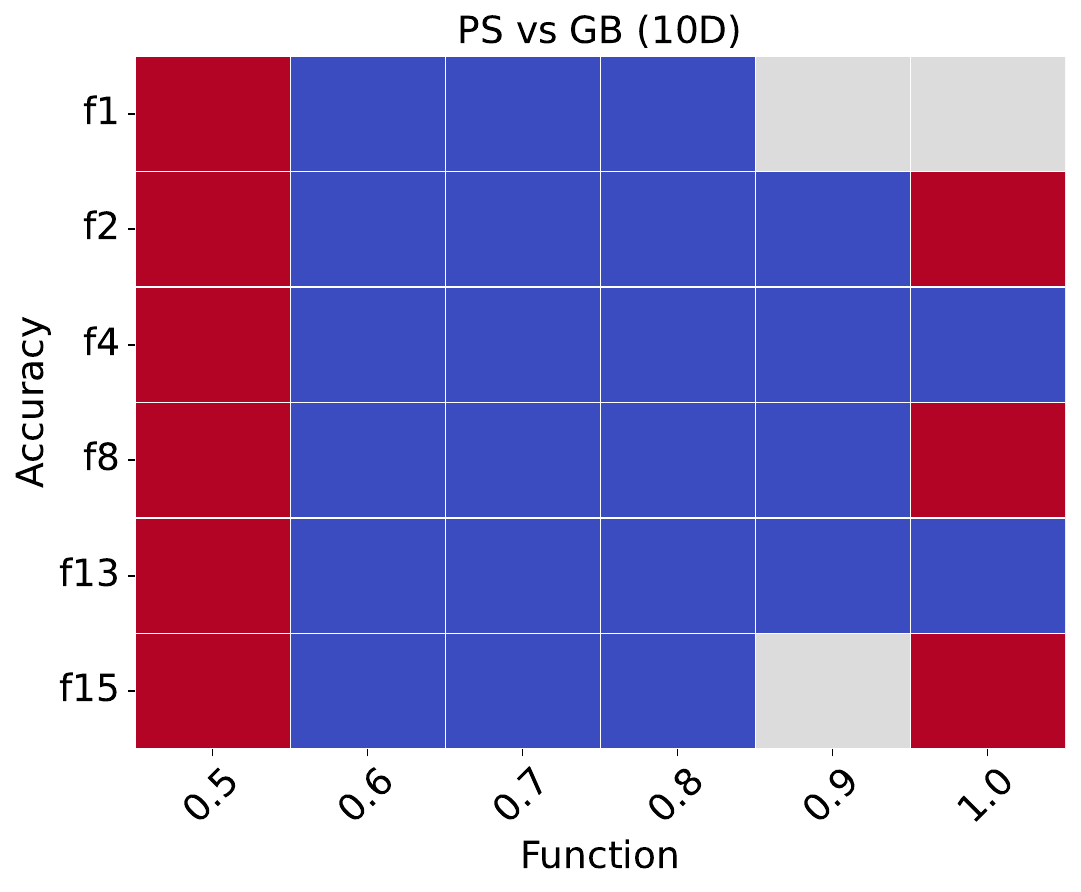}\label{fig:PS_vs_GB_10D}}
 \hfill
 \subfloat[IB vs GB on 10D]{\includegraphics[bb = 0 0 518 427, width=0.32\linewidth]{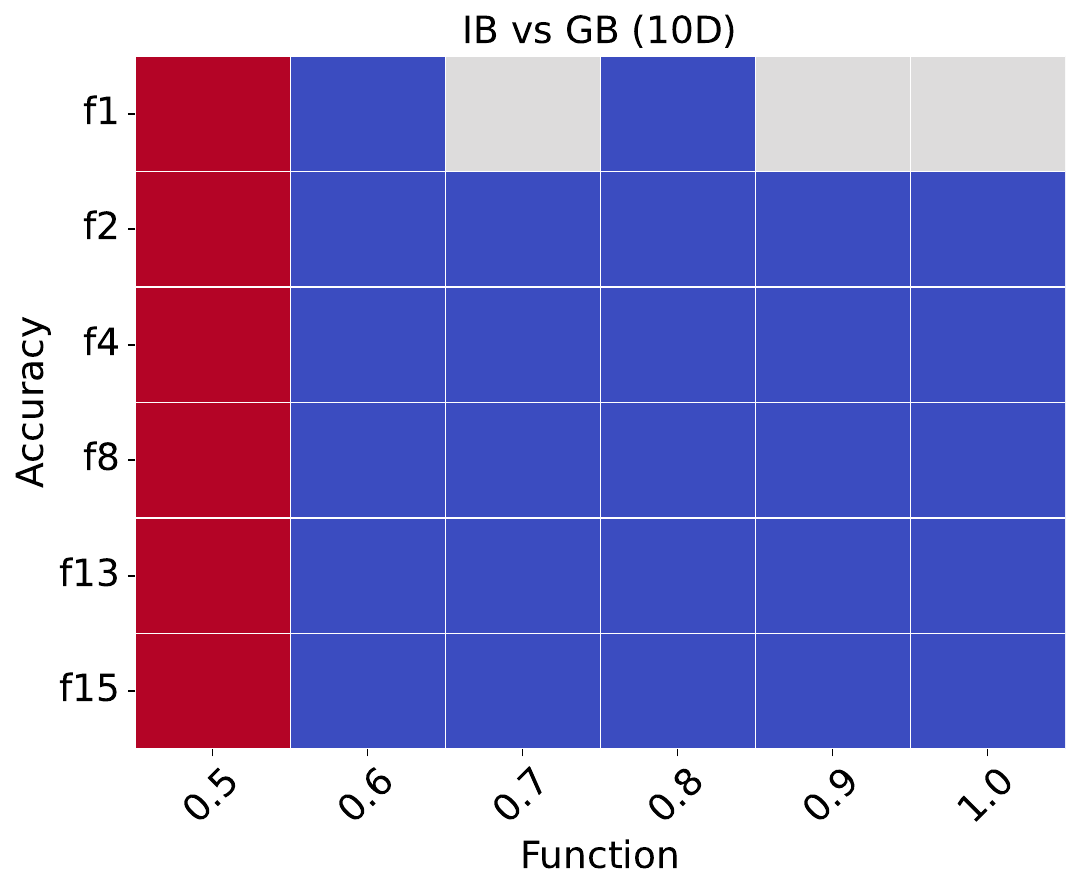}\label{fig:IB_vs_GB_10D}}
 \\
 \subfloat[PS vs IB on 30D]{\includegraphics[bb = 0 0 518 427, width=0.32\linewidth]{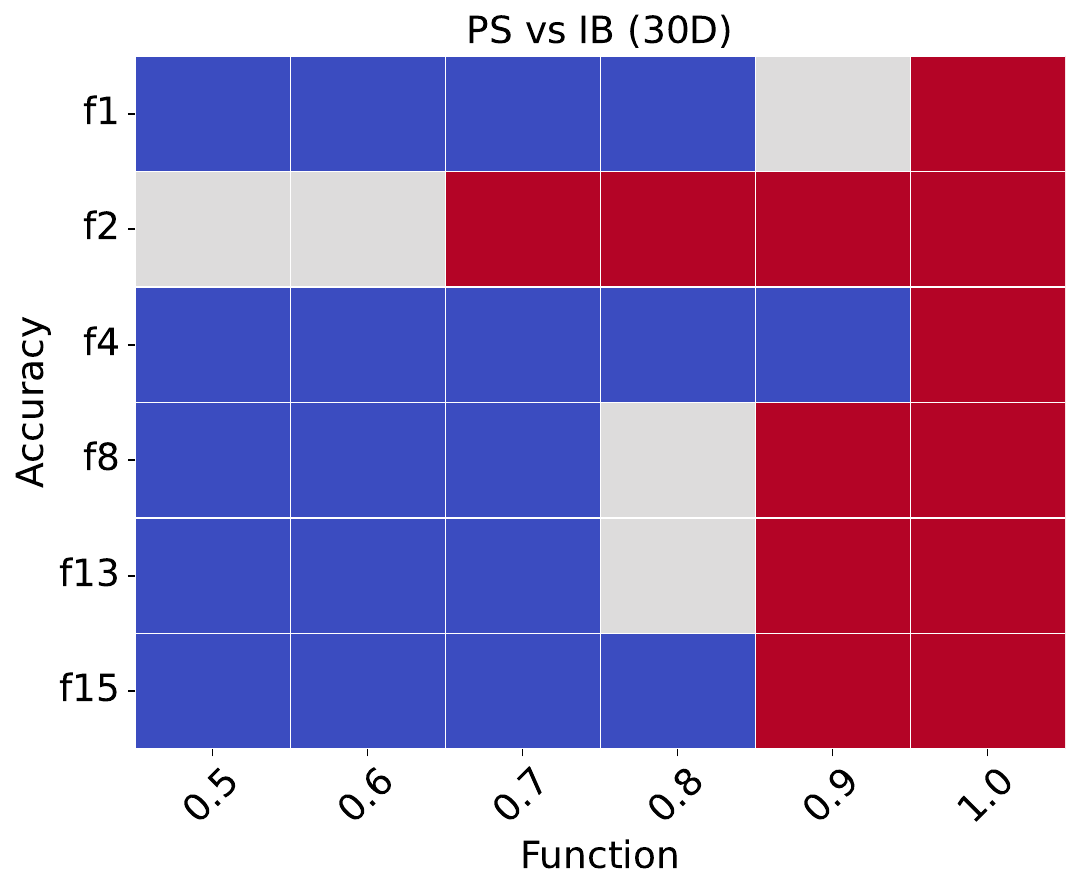}\label{fig:PS_vs_IB_30D}}
 \hfill
 \subfloat[PS vs GB on 30D]{\includegraphics[bb = 0 0 518 427, width=0.32\linewidth]{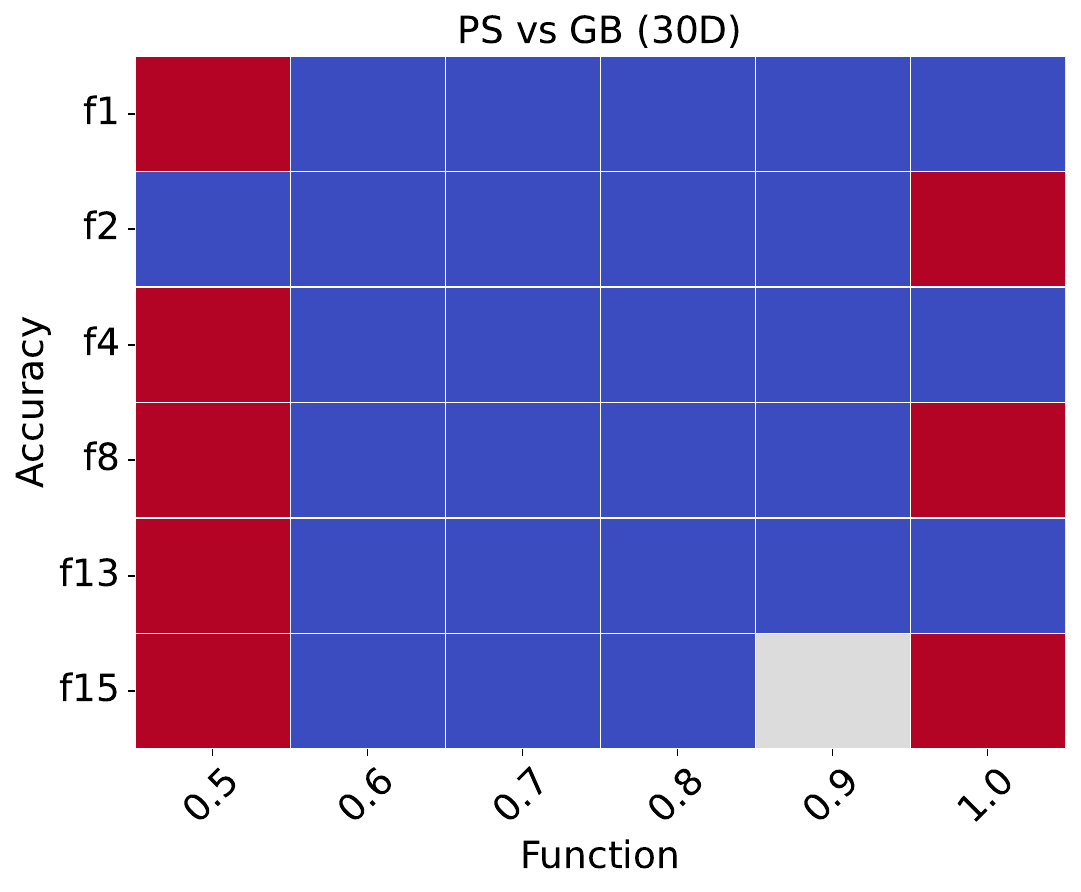}\label{fig:PS_vs_GB_30D}}
 \hfill
 \subfloat[IB vs GB on 30D]{\includegraphics[bb = 0 0 518 427, width=0.32\linewidth]{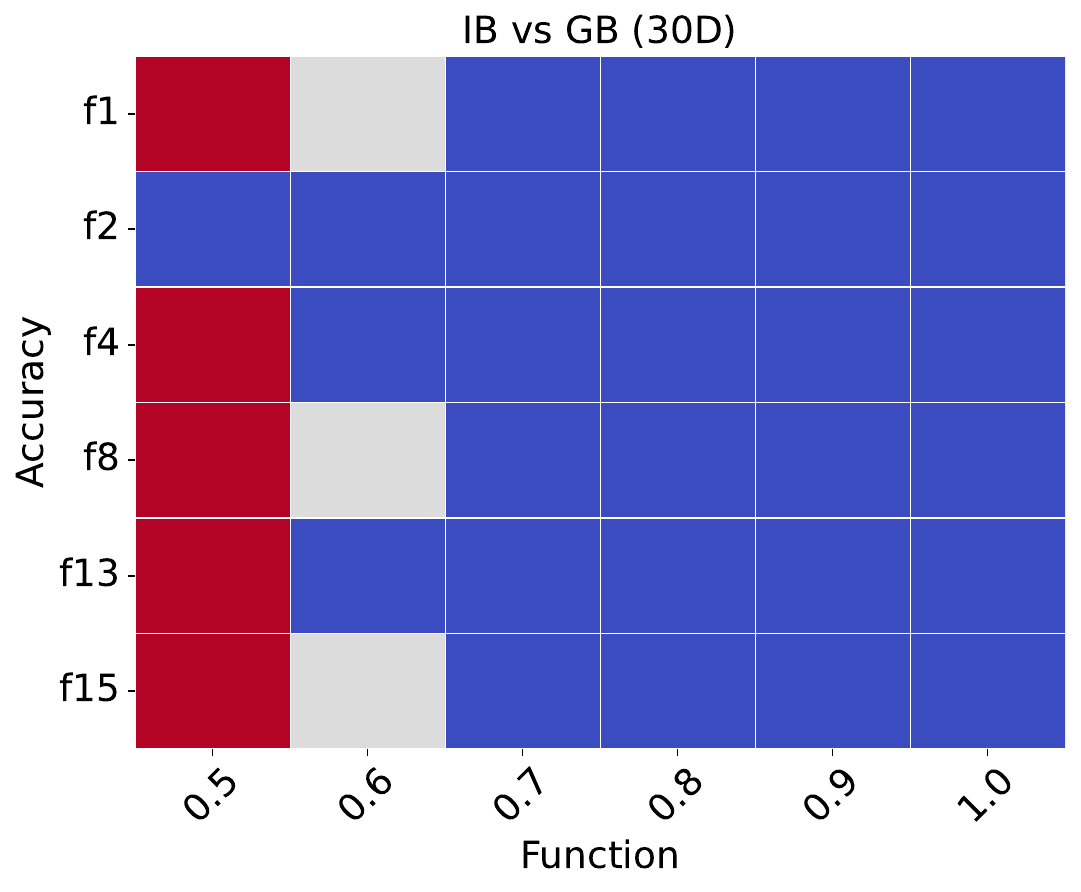}\label{fig:IB_vs_GB_30D}}
 \caption{Performance comparisons of PS, IB, and GB across the same accuracy using the Mann-Whitney U-test}
 \label{mannwhiteneyu}
\end{figure*}
Figure~\ref{mannwhiteneyu} shows the comparison results for each pair of model-management strategies. For the A versus B comparisons, red indicates that A has a better search performance than B, blue indicates worse performance, and gray indicates no significant difference.

For the comparison of PS and IB, Figures~\ref{fig:PS_vs_IB_10D} and \ref{fig:PS_vs_IB_30D} show that IB outperformed PS in both 10 and 30 dimensions when accuracy was low, whereas PS tended to perform better at high accuracy. This suggested that PS with a low-accuracy surrogate rejects offspring individuals that were better than their parents and should be selected for the next generation based on surrogate prediction, which negatively affected search performance.

For the comparison of PS and GB, Figures~\ref{fig:PS_vs_GB_10D} and \ref{fig:PS_vs_GB_30D} show that when the accuracy was 0.5, PS outperformed GB in all cases, except for $f2$ in 30 dimensions. However, when the accuracy was between 0.6 and 0.9, the GB either performed better or showed no significant difference. When the accuracy was 1.0, PS showed better results for functions $f2$, $f8$, and $f15$, where GB was better or did not differ from other functions.

For the comparison of IB and GB, Figures~\ref{fig:IB_vs_GB_10D} and \ref{fig:IB_vs_GB_30D} show that IB outperformed GB in all cases, except for $f2$ in 30 dimensions, with an accuracy of 0.5. However, GB tended to perform better at an accuracy of 0.6 or higher.

The comparison of GB and other strategies with an accuracy of 0.5 showed that GB had significantly worse search performance. This was attributed to GB using surrogate evaluations more frequently than IB or PS. The repeated use of incorrect predictions misled the search direction of the GB, and the final promising solutions obtained were of lower quality. However, when the accuracy was 0.6 or higher, the increased number of generations enabled better search performance.

Based on these results, IB was superior when the accuracy was 0.5, whereas GB was superior at an accuracy of 0.6 or higher. However, when the accuracy reached 1.0, PS demonstrated superior performance.

In this experiment, it remained unclear exactly where the advantage shifted between the 0.5 and 0.6 accuracy and between the 0.9 and 1.0 accuracy. In addition, we conducted further experiments at 0.01 intervals between 0.5 and 0.6 and 0.9 and 1.0 accuracy, followed by significance testing.

\begin{figure}[tb]
 \centering
 \subfloat[IB vs GB on 30D\label{ib-gb-comparison}]{
 \includegraphics[bb = 0 0 720 432, width=0.47\textwidth]{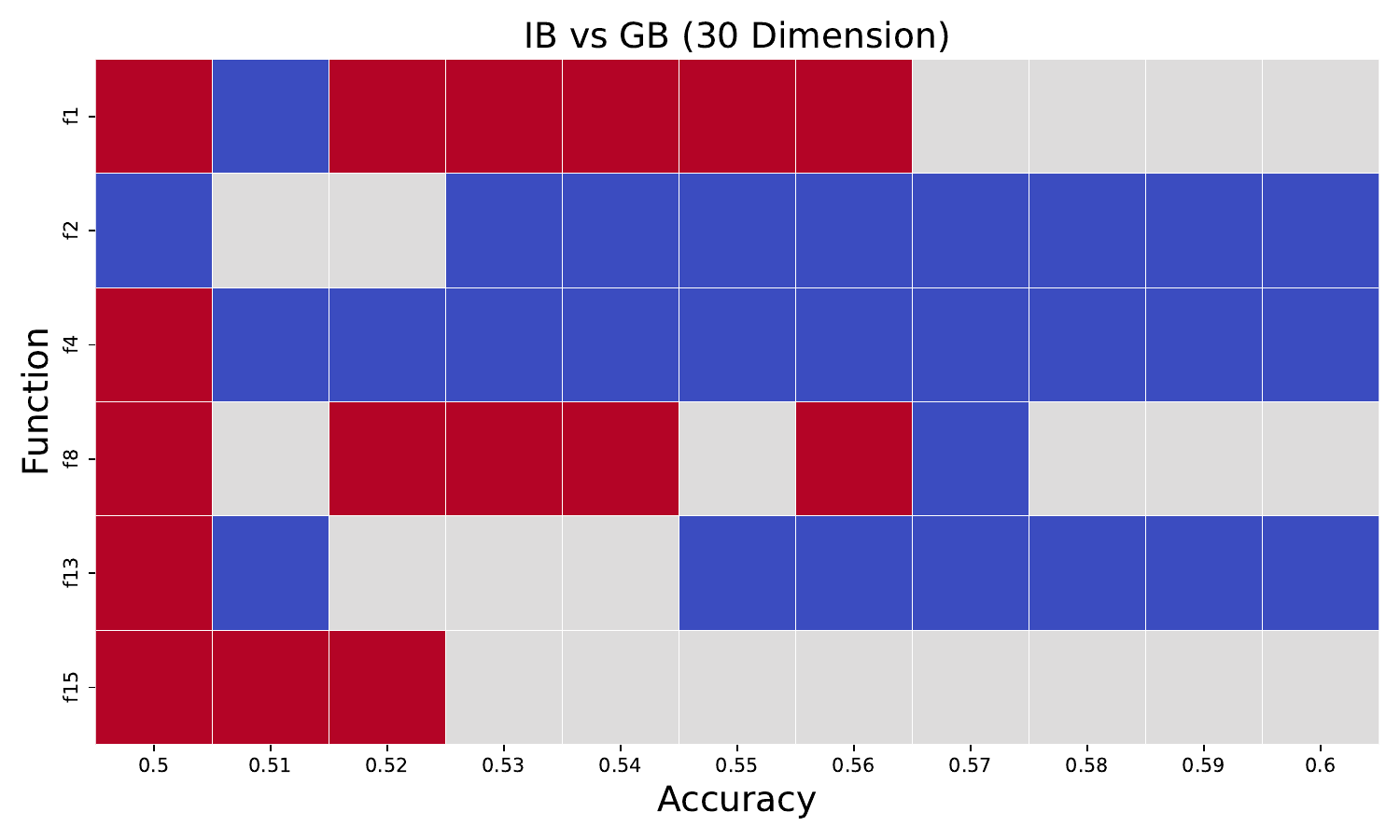}
 }
 \hfill 
 \subfloat[PS vs GB on 30D\label{ps-gb-comparison}]{
 \includegraphics[bb = 0 0 720 432, width=0.47\textwidth]{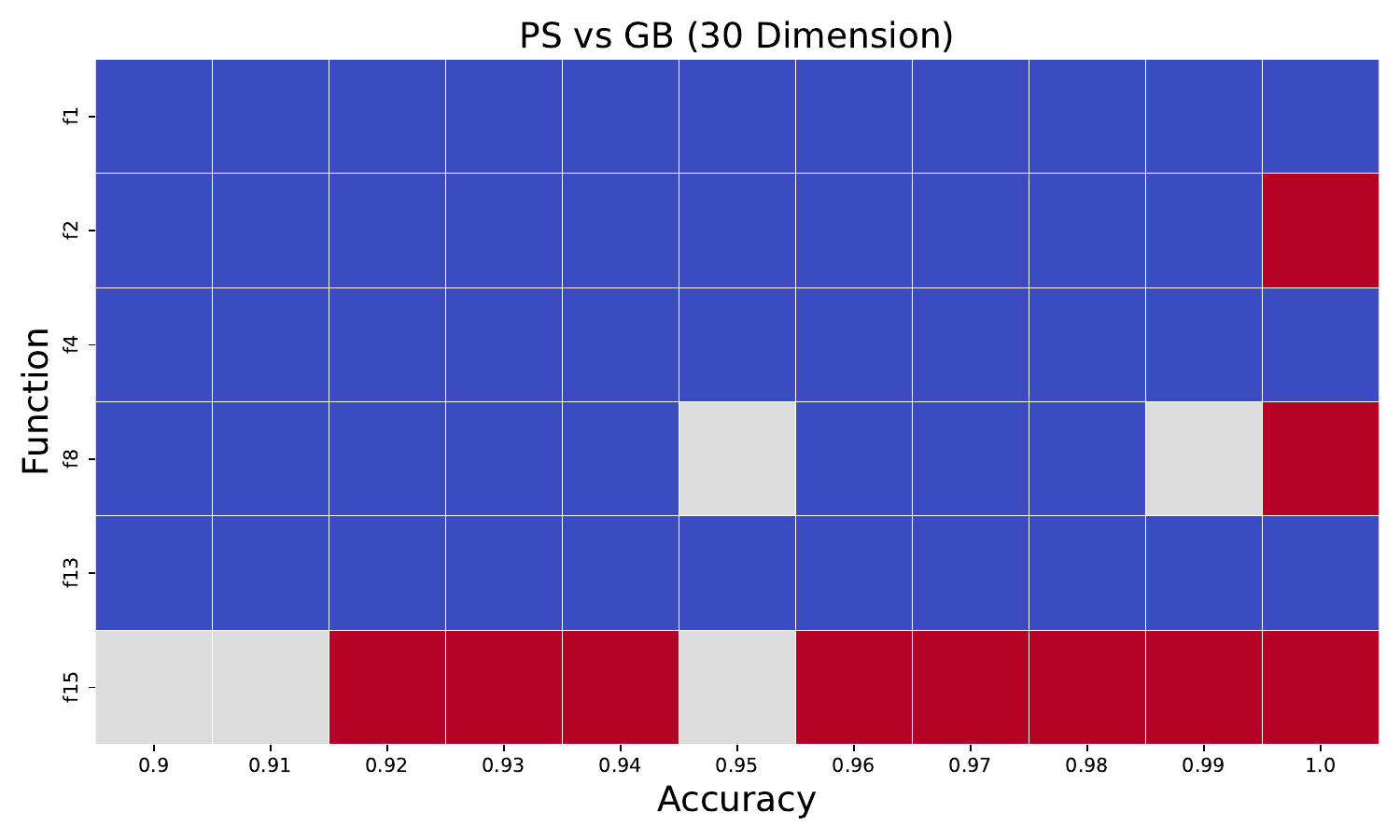}
 }
 \caption{Performance comparisons of IB and GB ($sp=[0.5, 0.6]$) and PS and GB ($sp=[0.9, 1.0]$) using the Mann-Whitney U-test}
 \label{fig:comparisons}
\end{figure}
The results of the Mann-Whitney U-test comparing IB and GB between accuracies of 0.5 and 0.6 (see Figure \ref{ib-gb-comparison}). In Figure \ref{ib-gb-comparison}, for the function $f1$, no significant difference was observed from an accuracy of 0.57 or higher. For $f2$, the GB became superior at 0.53 or higher. In $f4$, GB showed superiority from 0.51 or higher, and for $f8$, GB outperformed it at 0.57, but no significant differences were found thereafter. For $f13$, GB became superior with an accuracy of 0.55 and above, whereas for $f15$, no significant difference was observed at 0.53 or higher. Thus, GB should be selected when a surrogate accuracy of 0.57 or more could be ensured. Otherwise, IB was recommended when the accuracy was 0.56 or less.

Figure \ref{ps-gb-comparison} presents the statistical test results comparing PS and GB for surrogate accuracy levels ranging from 0.9 to 1.0. In Figure \ref{ps-gb-comparison}, for $f1$, GB consistently outperformed it; for $f2$, PS became superior only at an accuracy of 1.0. For $f4$, GB was always superior, whereas for $f8$, no significant difference was observed at 0.95 and 0.99, and PS became superior at 1.0. For $f13$, GB consistently outperformed, whereas for $f15$, no significant difference was found in the accuracies of 0.90, 0.91, and 0.95, with PS outperforming the other methods. Therefore, except in the case of $f15$, GB should be used unless an exact accuracy of 1.0 could be guaranteed.

Based on the above results, we could recommend guidelines for model management strategy selection.
\begin{itemize}
 \item Use IB when the accuracy was between 0.5 and 0.56
 \item Use GB when the accuracy was between 0.57 and 0.99
 \item Use PS when the accuracy was 1.0
\end{itemize}

\section{Conclusion}
This study aimed to analyze the impact of surrogate model prediction accuracy on SAEA search performance and its model management strategy. To this end, we constructed a pseudo-surrogate model that could set an arbitrary accuracy using actual evaluation function values.
To answer our research questions, we conducted experiments comparing three SAEA strategies (PS, IB, and GB) and pseudo-surrogate models with various prediction accuracies ranging from 0.5 to 1.0. We used six CEC2015 benchmark problems (unimodal, multimodal, and their composites) with 10 and 30 dimensions and analyzed the progression of minimum values discovered during the search process with a maximum of 2,000 actual evaluations.

For RQ1, we found that increasing the accuracy of surrogate models improved search performance, but the impact varied depending on the search strategy (PS, IB, or GB).

Specifically, for PS, the experimental results showed a better search performance with a higher prediction accuracy. When the accuracy was 0.5, the performance was almost equivalent to that without the surrogate. However, when the accuracy was 0.6 or higher, they outperformed the results without a surrogate, and using more accurate models appeared beneficial in PS.
For IB, positive correlations between accuracy and search performance were observed in all experimental results. However, the best performance was not always achieved when the accuracy was 1.0, though a certain level of accuracy was sufficient. In particular, for $f15$, the search performance decreased when the accuracy reached 1.0.
For GB, while searches with an accuracy of 0.5 and NoS showed little progress, significant performance improvements were observed with an accuracy of 0.6 or higher. Moreover, except for $f4$, there were no significant differences in the search performance for accuracies of 0.8 or higher.

Regarding RQ2, we confirmed the common tendency that a higher surrogate model accuracy enhanced search performance and that its impact varied depending on the model management strategy. 

In PS, no significant difference in search performance was observed between NoS and a surrogate model accuracy of 0.5, whereas search performance was improved by increasing surrogate model accuracy. In IB, differences were observed between the lower and higher accuracies. However, there was no significant difference in performance across the accuracies of 0.7 or higher. This suggested that the performance of IB stabilized beyond a certain accuracy threshold. Similarly, in GB, significant differences were observed between lower and higher accuracies, but no significant differences were found across accuracies of 0.8 or higher, suggesting that further accuracy improvements contributed minimally to the search performance beyond this threshold.

Finally, for RQ3, when the surrogate prediction accuracy was 0.56 or lower, it was preferable to use IB, whereas GB was recommended for accuracies of 0.57 or higher. However, if an exact accuracy of 1.0 could be ensured, using PS could be advantageous.

In our analysis, the impact of surrogate prediction accuracy on search performance and model management strategies in SAEA was examined under the assumption that the surrogate model’s accuracy remained constant and uniformly distributed across the search space. However, in practical SAEA scenarios, the accuracy tended to improve with generational progress, and local accuracy variations arose because of biases in the training samples. These factors were not replicated in the pseudo-surrogate model, and future studies should address these limitations. Other future work includes analyzing the relationship between the prediction accuracy and search performance in SAEAs requiring multiple surrogate models (e.g., constrained optimization problems and multi-objective optimization problems). In addition, we explored methods for switching search strategies during the optimization process based on the recommendations obtained in this study.

\section*{CRediT authorship contribution statement}
\textbf{Yuki Hanawa: } Writing -- original draft, Investigation, Methodology, Formal analysis, Software, Visualization. \textbf{Tomohiro Harada: } Writing -- review \& editing, Conceptualization, Funding acquisition, Methodology, Project administration. \textbf{Yukiya Miura: } Resources, Supervision.
\section*{Acknowledgments}
This research was supported by a Japan Society for the Promotion of Science Grant-in-Aid for Young Scientists (Grant Number 21K17826).
\section*{Declaration of competing interest}
The authors declare that they have no competing financial interests or personal relationships that could influence the work reported in this study.
\section*{Data availability}
No relevant data were used in the research described in this study.
\section*{Declaration of generative AI and AI-assisted technologies}
During the preparation of this work the authors used ChatGPT in order to improve language and readability. After using this tool/service, the authors reviewed and edited the content as needed and take full responsibility for the content of the publication.









\bibliographystyle{elsarticle-harv} 
\bibliography{chde,
citation,
de,
citation-347650551,
constraint-handling_in_nature-,
IEEE-Xplore-Citation,
name,
recent-advances-in-surrogate-based-optimization,
surrogate_models_in_evolutionary_single_,
Surrogate-assisted_evolutionary_computation,
jin2005,
generation_pso.bib,
S0020025520311609,
S0020025524001592.bib,
references.bib},

\end{document}